\documentclass[runningheads]{llncs}
\usepackage[T1]{fontenc}
\usepackage{graphicx}
\usepackage{booktabs}
\usepackage[misc]{ifsym}

\usepackage{microtype} % helps \looseness=-1 a lot, I hope this is legal :D

\usepackage{booktabs}
\usepackage{tabularx}
\usepackage{color}
\usepackage{amsmath,amssymb,amstext}
\usepackage{multirow}
\usepackage{bbm}
\usepackage{bm}
\usepackage{wrapfig}

\usepackage{mathtools}

\usepackage{algorithm}
\usepackage{algorithmic}
\DeclarePairedDelimiterX{\infdivx}[2]{(}{)}{#1\;\delimsize\|\;#2}

\usepackage{enumitem}

\usepackage{multirow}
\begin{document}

\title{Multi-Depth Concept Extraction for Post-Hoc Vision Encoder Explanation}

\titlerunning{Multi-Depth Concept Extraction for Post-Hoc Vision Encoder Explanation}

%\author{Ahcène~Boubekki\inst{1}\orcidID{0000-0003-1606-1513} \Letter \and
%Samuel~G.~Fadel\inst{2}\orcidID{0000-0002-4459-4336} \and
%Sebastian~Mair\inst{3}\orcidID{0000-0003-2949-8781}}

\author{Ahcène~Boubekki\inst{1} \Letter \and
	Samuel~G.~Fadel\inst{2} \and
	Sebastian~Mair\inst{3}}

\authorrunning{A. Boubekki et al.}

\institute{University of Copenhagen \email{ahcene.boubekki@math.ku.dk}
\and
Technical University of Denmark \email{samma@dtu.dk}
\and
Linköping University
\email{sebastian.mair@liu.se}}

\toctitle{Multi-Depth Concept Extraction for Post-Hoc Vision Encoder Explanation}

\tocauthor{Ahcène~Boubekki, Samuel~G.~Fadel, Sebastian~Mair}

\maketitle

\begin{abstract}
Explainable AI methods for vision models aim to identify the parts of the input that are important for the final prediction and subsequently relate these regions to human-understandable concepts. Here, we propose focusing solely on the encoder and relating its intermediate outputs to the input, instead. We introduce Neuro-Activated Vision Explanations (NAVE), a post-hoc, unsupervised, and architecture-agnostic (across CNNs and ViTs) method for extracting and visualizing internal representations from frozen vision model encoders. Specifically, NAVE clusters composite feature activations from multiple encoder depths to produce interpretable segmentation maps with controllable granularity, requiring no fine-tuning or architectural modifications. Through extensive experiments, we quantitatively demonstrate that NAVE's concepts align with input semantics and can be used in downstream tasks. We further demonstrate NAVE as an inspection tool by analyzing how training strategies and architectures affect encoder representations. Overall, our results establish NAVE as an effective tool for post-hoc model inspection and enhancing transparency in vision models.
\texttt{\url{https://github.com/Ahcene-B/NAVE}}

\end{abstract}

\section{Introduction}

The adoption of deep learning in safety-critical domains, including medical diagnostics~\cite{borys2023explainable} and autonomous surveillance~\cite{williford2020explainable,panfilova2024applying}, has triggered a regulatory response that has accelerated the development of Explainable AI (XAI).
Therefore, understanding the internal mechanisms of Convolutional Neural Networks (CNNs) and Vision Transformers (ViTs)~\cite{ViT}, the standard backbones in vision applications, is essential. Although significant research has focused on explaining a model’s \emph{predictions}, comparatively little attention has been devoted to understanding a model’s \emph{perceptions}. This disparity likely results from the lack of an effective and efficient representation of the information encoded in the encoder.\looseness=-1

Emblematic XAI methods, such as Integrated Gradients~\cite{IG} or Grad-CAM~\cite{GC}, aim to identify which input pixels most influence a specific output class (e.g., ``Which pixels make this look like a dog?''). Their structural class-dependence means they provide only limited insights into parts of the input unrelated to any class. Moreover, inconsistencies among explanations of different methods undermine trust and reliability, thus limiting their usage beyond their original purpose. Another line of research attempts to relate predictions to human-understandable concepts, but defining or extracting these concepts remains a significant challenge. Indeed, predefined dictionaries~\cite{kim2018interpretability} or \emph{ad-hoc} segmenters~\cite{li2021instance} may not align with the model’s internal mechanisms. Jointly learning concepts with predictions necessitates architectural modifications, thereby explaining another model.\looseness=-1

The question we aim to answer here is: ``What does the encoder capture prior to considering a label?'' Indeed, understanding these internal representations, or decoding the encoder’s \emph{perception}, is essential for verifying, inspecting, and debugging models. 
Because of this, we take a \emph{post-hoc} stand and acknowledge the different roles the encoder and classifier play. We discuss the extraction of concepts from a frozen encoder and treat the explanation of the prediction as a downstream task.
This question arose early in the onset of deep learning and of CNNs, and was initially studied from the perspective of multilayer perceptrons. At the time, the intuition was that each neuron carries specific information~\cite{zeiler2014visualizing}, as in Net2Vec~\cite{net2vec}, which nevertheless concluded that different neurons share the same information. 
Caron et al.~\cite{DINO} used a Principal Component Analysis (PCA) to demonstrate how DINO ViTs can distinguish primary objects from the background and exhibit a sense of depth. The analysis has been refined~\cite{RabbitHull} by extracting a very large number of concepts using a sparse autoencoder. However, the visualizations still depend on three-directional PCA, which is insufficient for inspecting and representing complex scenes. Similar in spirit to our proposal are the works of Chormai et al.~\cite{chormai2024disentangled} and Jacob et al.~\cite{jacob}, which use Layer-wise Relevance Propagation (LRP)~\cite{LRP} to visualize concepts represented by the directions in the factorization of an intermediate embedding. While these methods facilitate inspection of internal mechanisms and yield novel mechanistic insights, they are fundamentally constrained by the complexity of their protocols.\looseness=-1

We introduce Neuro-Activated Vision Explanations (NAVE), an unsupervised method for revealing and visualizing the internal semantic maps a frozen vision encoder makes of an input. NAVE extracts concepts by clustering, using $k$-means, composite pixels from feature activations drawn from multiple encoder depths. The method is simple, efficient, and easy to understand. The returned explanation maps is a concept-based segmentation map~\cite{cho2021picie} revealing the semantical spatial organization extracted by the embeddings. The explanation can be refined by controlling the number of clusters. NAVE is not dependent on the classifier's predictions or the label, meaning that retraining only the classifier does not affect the explanations.
NAVE provides a ``lens'' into the encoder, enabling practitioners to:\looseness=-1
\begin{itemize}
    \item[(i)] Visualize internal concepts through unsupervised segmentations of instances derived from the encoder’s internal representations (Section~\ref{sec:method}).
    \item[(ii)] Quantitatively evaluate through weakly-supervised object localization, the alignment of the concepts with the dataset semantics (Section~\ref{sec:eval}).
    \item[(iii)] Inspect and debug models by identifying clues of poor training robustness, artifacts, and shortcut saturation in NAVE’s explanation maps (Section~\ref{sec:inspect}).
\end{itemize}

By shifting the focus from attribution to representation and providing efficient visualizations, NAVE facilitates a more comprehensive mechanistic understanding of vision models. It serves as a tool for both post-hoc inspection and the advancement of more transparent and reliable AI systems.\looseness=-1

\section{Related Work}

The inspection of feature activations from CNNs has been the focus of two major works. DeepDream and related work~\cite{DeepDream,erhan2009visualizing} offer a visual inspection of what causes the activation of each neuron, confirming that deeper layers capture higher-level concepts. However, the instability of the underlying activation maximization limited its reach. A second breakthrough is presented in~\cite{DINO}, which shows that the self-attention of the class token of self-supervised ViTs carries semantic information about the input. This work was pursued in~\cite{DINOv2}, where field-of-depth is also recovered from attention and activations using a PCA. The method does not, however, apply to non-ViT CNNs, and adding more PCA components does not really refine the decomposition. More recently, \cite{RabbitHull} extracts a large concept inventory from DINO representations using a sparse autoencoder, but visualizations still depend on three-directional PCA projections.
Hypercolumns~\cite{hariharan2015hypercolumns} introduced per-pixel descriptors by concatenating multi-layer CNN activations for supervised segmentation tasks. NAVE repurposes this aggregation idea in a post-hoc, unsupervised XAI setting: rather than serving a labeled downstream task, the composite pixels are clustered to reveal what the encoder retrieves prior to any prediction.
Besides, NAVE applies to any CNN-based architecture, including ViT, and the granularity of its explanation maps can be controlled.
\looseness=-1

Concept-based models propose to leverage a set of concepts to provide human-friendly explanations~\cite{kim2018interpretability}. While CRAFT~\cite{fel2023craft} and Net2Vec~\cite{net2vec} extract concepts from the final encoder block, CONE-SHAP~\cite{li2021instance}, EAC~\cite{sun2023explain}, and VCC~\cite{VCC} link the concepts to regions of the input obtained through an ad-hoc segmenter (SLIC~\cite{SLIC}, SAM~\cite{SAM}). Relying on such segmenters means the resulting decomposition does not reflect what the model learned during training. Besides, applying the encoder to masked patches likely falls out of distribution at each depth. NAVE instead derives its explanations directly from a combination of encoder activations, meaning it is sensitive to the model's training.

Self-explainable models usually learn concepts either in the embedding space~\cite{alvarez2018towards} or in the activation space of the last block~\cite{protopnet}. The drawback of learning a transparent classifier with prototypes is that, although performance may match an opaque model, the learning is different and so are the learned concepts and their interactions~\cite{kmex}. NAVE takes a post-hoc stand on a frozen, unmodified encoder, ensuring that the extracted concepts reflect exactly what the original model has learned.
\looseness=-1

\begin{algorithm}[t]
    \caption{NAVE Python pseudocode. See the text for details on the notation.}
    \label{alg:nave}
    \begin{algorithmic}[1]
        \REQUIRE \texttt{X: Batch of dimension  (N, H, W),\\
         Selected\_Blocks: Ordered list of block ids,\\
         K: Number of clusters.}
        \texttt{
        \STATE min\_id = min(Selected\_Blocks)
        \STATE Auxiliary = empty((N $\times$ H$_\texttt{min\_id}$ $\times$ W$_\texttt{min\_id}$, 0))
        \STATE z = X
        \FOR {b = 0 to B} 
            \STATE z = f$_\texttt{b}$(z) 
            \IF{ b in Selected\_Blocks}
                \STATE \hspace{3pt}u = upsample(z, (H$_\texttt{min\_id}$, W$_\texttt{min\_id}$))
                \STATE \hspace{3pt}u = reshape(u, (N $\times$ H$_\texttt{min\_id}$ $\times$ W$_\texttt{min\_id}$, C${}_\texttt{b}$)) 
                \STATE \hspace{3pt}u = u / euclidean\_norm(u, axis=2)
                \STATE \hspace{3pt}u = u * C$_\texttt{min\_id}$ / (1+C$_\texttt{b}$)
                \STATE Auxiliary = concatenate(Auxiliary, u, axis=2)
            \ENDIF
        \ENDFOR
        \STATE Explanations = KMeans(K).fit\_predict(Auxiliary)
        \STATE Explanations = reshape(Explanations, (N, H$_\texttt{min\_id}$, W$_\texttt{min\_id}$))
        \RETURN Explanations
        }
    \end{algorithmic}
\end{algorithm}

\section{Neuro-Activated Vision Explanations}\label{sec:method}

A trained vision encoder $f$ consists of $B+1$ blocks, denoted as $f_0, f_1, \ldots, f_B$. The initial block $f_0$ typically comprises preprocessing layers, while subsequent blocks $f_b$s are repeated blocks specific to the architecture such as ResNets~\cite{ResNet} or ViTs. The network processes image batches with shape $(N, W, H, C)$, and each block outputs a tensor of shape $(N, W_b, H_b, C_b)$. NAVE operates on feature maps before the final average pooling, enabling spatial segmentation. Consequently, the final average pooling is not in $f_B$.
\looseness=-1

\textbf{Algorithm.}
Algorithm~\ref{alg:nave} presents a Python-style pseudocode for NAVE. The method requires two parameters: \texttt{Selected\_Blocks}, an ordered list of block indices for extracting intermediate outputs, and $K$, the number of clusters for grouping \textit{composite} pixels. The procedure begins by sequentially processing the batch input through each block of the network.
For each block included in $\texttt{Selected\_Blocks}$:
\begin{itemize}[topsep=0pt]
    \item The output of the current block is upsampled to match the spatial dimensions of the output from the shallowest selected block (\texttt{min\_id}).
    \item The output is normalized and scaled down by a factor proportional to the number of channels.
    \item The resulting tensor is concatenated to the auxiliary matrix \texttt{Auxiliary}.
\end{itemize}
After processing all blocks, $\texttt{Auxiliary}$ is clustered into $K$ groups using $k$-means. The centroids can also be referred to as the \textit{prototypes} of their cluster. The final \texttt{Explanations} correspond to the cluster assignments and are reshaped to match the output dimensions of block $f_{\texttt{min\_id}}$.\looseness=-1

Before clustering, the matrix \texttt{Auxiliary}, which aggregates the feature activations, has dimensions $(N\!\times\!H_\texttt{min\_id}\!\times\!W_\texttt{min\_id}, \sum_{b \in \texttt{Selected\_Blocks} } C_b)$. The spatial consistency of convolutional receptive fields implies that each row represents a \emph{composite} pixel approximately in the same location in the input. In a ViT, the receptive field remains constant because each pixel (or token) corresponds to a specific patch of the input. The final explanation map can therefore be interpreted as an unsupervised segmentation of the input, based on the cross-layer representations summarized in the composite pixels.\looseness=-1

The multi-depth design is the key to NAVE's explanatory power. Single-depth variants (e.g., N4) use only the most abstract representations, producing coarse maps that lose spatial detail. Earlier layers retain fine-grained edge information but lack semantic abstraction. By concatenating normalized activations across depths, NAVE produces composite pixels that simultaneously encode both local texture and global semantics, see Tables~\ref{tab:acc} and \ref{tab:wsol-r50}, and Figure~13 in Appendix~I. The scaling and normalization allows to balance the importance of the feature activation despite their different number of channels. See Appendix~D for an ablation study.

\begin{figure}[!ht]
	\centering
    \hspace{2.5cm} NAVE N34 \hfill NAVE N234\hspace*{1.9cm} \\
    \vspace{1em}
	\hfill
	\includegraphics[width=.4\linewidth]{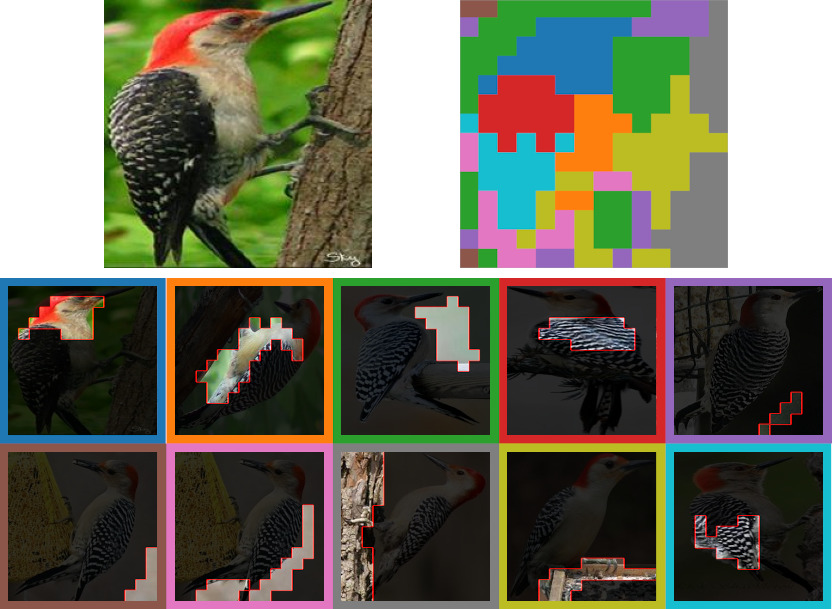}
	\hfill
	\includegraphics[width=.4\linewidth]{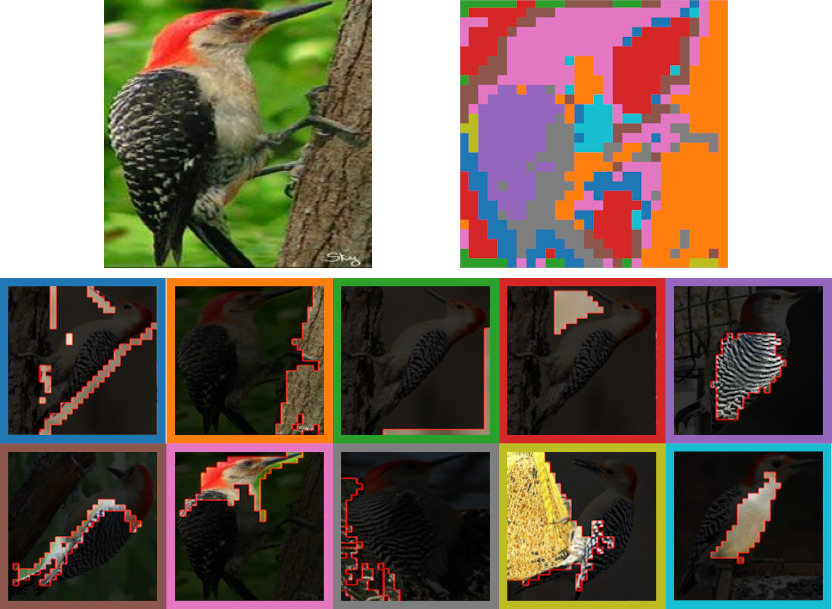}
	\hfill
	\caption{An illustration of the interpretation process of NAVE using different intermediate layers of a ResNet50 pretrained on ImageNet1K. From left to right and top to bottom: input image of a red-bellied woodpecker from the CUB200 dataset; explanation map computed using the feature activations of indicated blocks. Below, the closest patches in the training set for each class prototype (centroid) are shown. The border color of the patches corresponds to the segments of the explanation maps.}
	\label{fig:interp}
\end{figure}

\textbf{Discussion.}
The method requires three inputs: a batch of images, the number of clusters, and the indices of the blocks from which feature activations are extracted.
Although the method can be applied to a single image, it is recommended to pretrain the internal clustering on a set of images from the same class. This strategy ensures that the cluster semantics remain consistent across examples within the same class. Representative patches can be used to interpret the clusters (see Figure~\ref{fig:interp}). Unless otherwise specified, in all examples, one $k$-means is pretrained on each class of the training set, and all centroids are subsequently merged into a single algorithm.

The number of clusters and selected blocks can be determined using the elbow method or the requirements of a downstream task (see Section~\ref{sec:acc}). Selecting blocks closer to the input or output involves a trade-off that affects downstream analyses. Shallower layers produce explanation maps that align more closely with input edges, while deeper layers yield more abstract representations that reflect the classifier’s information. Empirical results indicate that the last three blocks provide the most informative explanations for ResNets smaller than ResNet34, the last two for ResNet50\footnote{The ResNet50 is dilated at the last block.} and larger, and the output of the last transformer block for ViTs. We introduce the notation Nxyz (or N:x-y-z if depth exceeds 9), where x, y, z denote the selected block indices, e.g., N234 for ResNet34, N34 for ResNet50, and N:12 for a ViT with 12 transformer blocks.
\looseness=-1

\textbf{Limitations.}
By default, $k$-means clustering is employed due to its computational efficiency and simplicity. However, it exhibits poor scalability with respect to the number of composite pixels $N\times H_\texttt{min\_id}\times W_\texttt{min\_id}$ and produces stochastic outputs due to the initialization. Scalability can be improved by using the mini-batch variant~\cite{minikm}. Hierarchical clustering~\cite{gordon1987review} can provide deterministic explanations, but its computational cost increases quadratically with the number of pixels.\looseness=-1

\textbf{Interpretation Process.}
Figure~\ref{fig:interp} illustrates the interpretation process of NAVE for a ResNet50 pretrained on ImageNet1K. The explanation maps identify regions of the input that produce similar feature activations. Consequently, spatially disconnected segments may be assigned to the same cluster, represented by its color. Cluster interpretation is supported by visualizing the segment closest to each associated prototype (the $k$-means centroid) using a consistent color scheme. The original scene is more discernible in the explanation generated by N234 than in that produced by N34, indicating a stronger information compression in the third block. Incorporating shallower feature activations enhances the resolution and detail of the explanation map. In contrast, coarser explanations are advantageous for automatic object localization, as discussed in Section~\ref{sec:wsol}.
\looseness=-1

\subsection{Comparing PCA and $k$-means}

\begin{figure}[!ht]
    \centering
    \includegraphics[width=\linewidth]{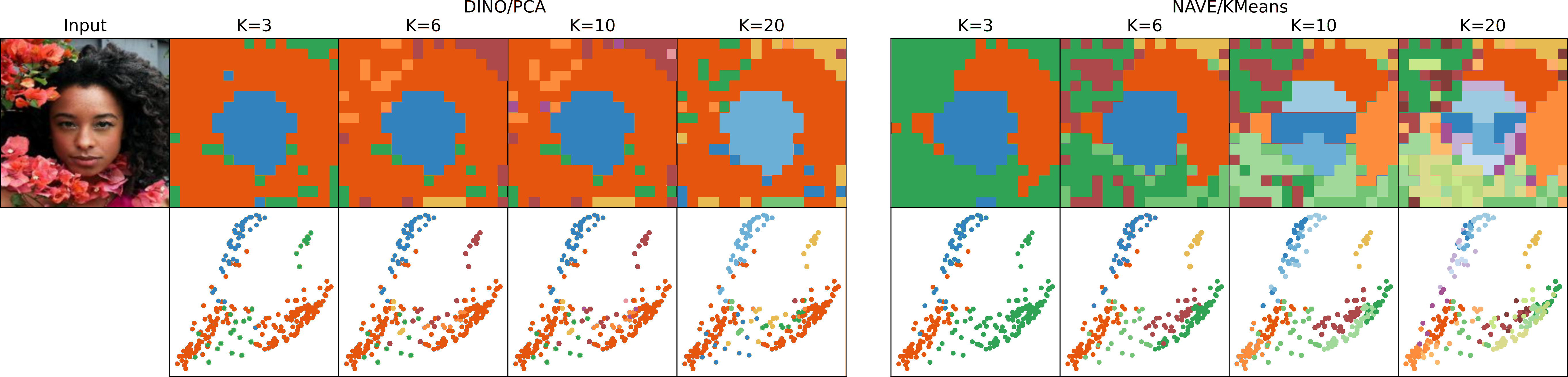}
    \caption{Explanations derived from DINO/PCA and NAVE/$k$-means applied to a ViT-S/14 with increasing number of components/clusters ($K$). While NAVE uses the extra clusters to refine the decomposition of the scene, the PCA-based one barely changes beyond $6$ components.
    \looseness=-1
    }
    \label{fig:pca_km}
\end{figure}

The original DINO paper~\cite{DINO} introduced a visualization method based on three-dimensional principal component analysis (PCA) to examine how the model interprets the scene. However, a limitation of PCA is that adding more components primarily refines the last components. Figure~\ref{fig:pca_km} illustrates this phenomenon by comparing DINO/PCA and NAVE/$k$-means explanations for an increasing number of components or clusters ($K$). Both methods are applied to the output of the final transformer block (N:12) of a ViT-S/14 model trained on DINOv2, using an input image from the CelebA dataset. In the PCA segmentations, pixels are assigned to the component with the greatest absolute value. The same color scheme is maintained across plots for the PCA components, with the first three directions in red, blue, and green, respectively. The corresponding partition in the token embedding is shown below each explanation map. Additional examples are provided in Appendix~B. 

For $K=3$, PCA explanations isolate portions of the main objects. When the number of components increases to $6$, parts of the first PCA direction is diluted into new ones. Overall, the segmentation remains stable across all settings, with only the face correctly isolated, while the hair is merged with the flowers. In contrast, NAVE explanations provide increasingly detailed scene interpretations as the number of clusters increases. Unlike PCA, $k$-means with $K=3$ accurately distinguishes the hair from the flowers. Although object color contributes to cluster assignments, the nature and context of the objects also play a significant role. For example, at $K=10$, the upper and lower flowers are assigned to different clusters. In summary, while PCA performs adequately with three components, $k$-means offers greater flexibility and accuracy for recovering and interpreting scene layouts from the network output.

\subsection{Qualitative Comparison}

\begin{figure}[!ht]
	\centering
	\includegraphics[width=.9\linewidth]{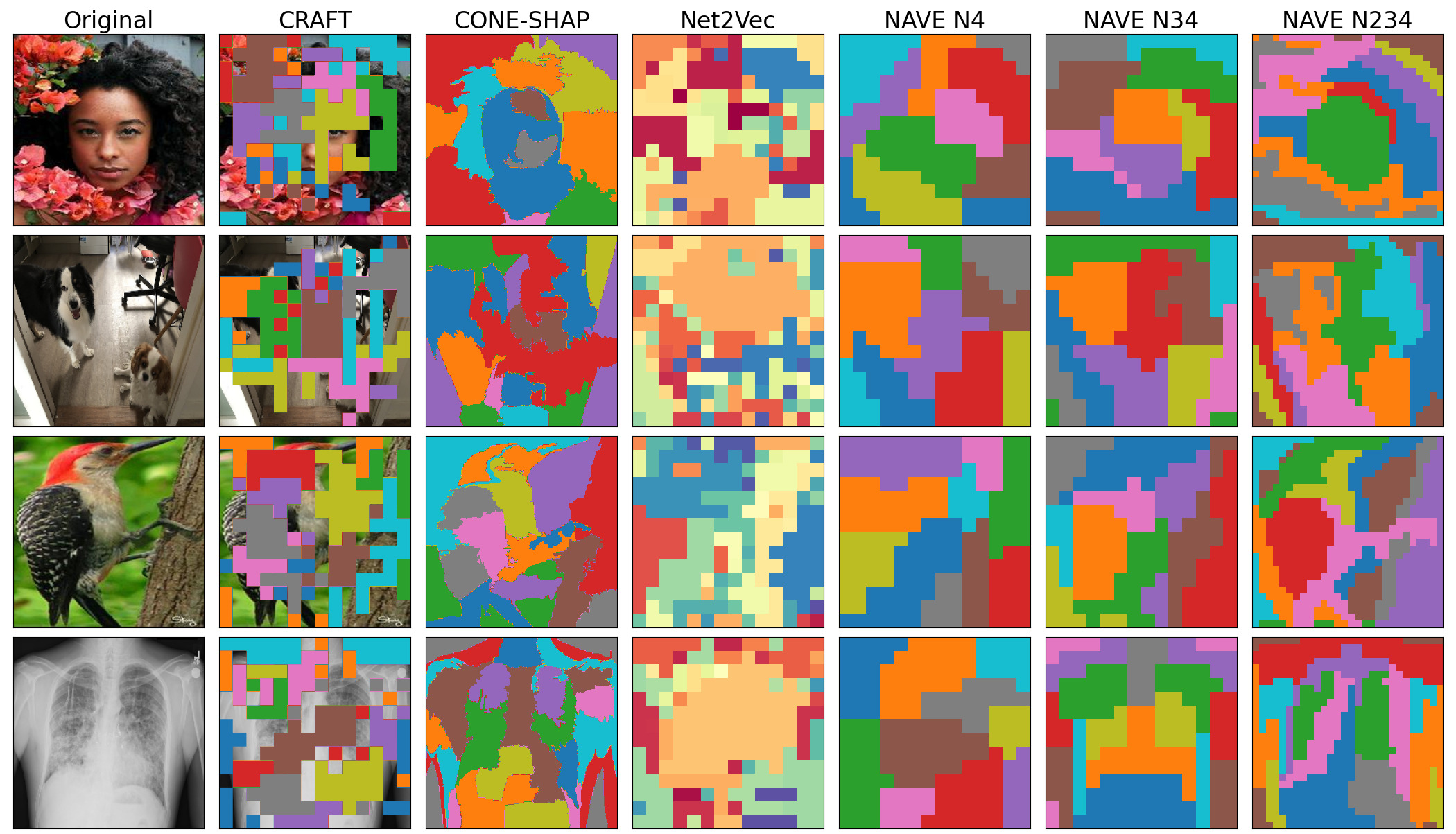}
	\caption{Explanation maps computed image-wise (K = 10) for three baselines and three NAVE configurations applied to a ResNet50 on examples taken from the CelebA, VOC07, CUB200 and Chest-X-Ray datasets. For Net2Vec, the map corresponds to the argmax over the 2048 output dimensions of N4.}
	\label{fig:expl}
\end{figure}

Figure~\ref{fig:expl} presents explanation maps for a ResNet50 generated per image (batch size of 1, $K=10$) by three baselines and three NAVE configurations, using images from diverse datasets.
Each method is evaluated according to two criteria: the recognizability of the scene organization and the extent to which spatially disconnected yet visually similar regions are assigned to the same cluster.
\looseness=-1

Among the baseline methods, CRAFT produces sparse maps that resemble feature-attribution maps, with scattered regions covering the input. CONE-SHAP explanations resemble more to a SLIC superpixel segmentation rather than the model's internal representation. In the bird image, the background is correctly grouped despite spatial discontinuities, indicating that semantically similar regions can be clustered together. However, the quality of the explanation is constrained by the ad hoc segmentation; for example, the two dogs are not visible in the second example. Net2Vec maps, computed as the argmax over the 2048 dimensions of the N4 output, broadly capture the global organization of the original scene, but the large number of clusters makes the maps difficult to interpret. For the NAVE configurations, the impact of incorporating shallower blocks is evident. N4, which operates solely on the final block, produces coarse maps. As shallower blocks are progressively included, the explanation maps become more detailed and interpretable. For N234, the scene is clearly recognizable, and spatially disconnected yet semantically related regions are consistently clustered together. NAVE's explanations of the woman portrait differ from those in Figure~\ref{fig:pca_km} because the model to explain is here a ResNet50 instead of a ViT-S/14. In the bird image, scattered background regions are assigned to the same concept, while in the X-ray, the two lungs, though mirrored and spatially disconnected, are clustered together. These observations illustrate the benefits of NAVE's multi-depth explanation maps.
\looseness=-1

\section{Evaluations}\label{sec:eval}

Having established qualitatively the advantages of multi-depth aggregation and $k$-means decomposition, we evaluate now NAVE quantitatively along two axes: whether its concept representations carry sufficient information for classification, and whether they align with the semantic content of the input. We extend the latter by an analysis of the sensitivity of NAVE to training settings.

\subsection{Experimental Setup}

\textbf{Datasets.}
For a proxy object localization task, we use VOC07 (training + validation subsets)~\cite{pascal-voc-2007}, VOC12~(training + validation subsets)~\cite{pascal-voc-2012}, and COCO20k (training subset)~\cite{lin2014microsoft}.
Furthermore, we use Chest-X-RAY~\cite{arias2024analysis}, STL-10~\cite{coates2011analysis}, and ImageNet1K~\cite{russakovsky2015imagenet} to evaluate the influence of the data used to train the models.
Other examples are extracted from the CelebA dataset~\cite{celeba}.
\textbf{Implementation.}
We use ViT-Small (ViT-S) and ViT-Base (ViT-B) with patch sizes of $14\times14$ and $16\times16$ \cite{ViT}, and residual networks, i.e., ResNet18 and ResNet50~\cite{ResNet} using implementations and pretrained weights from PyTorch~\cite{pytorch} or DINO's repositories~\cite{DINO,DINOv2}.
\textbf{Baselines.}
For concept explanation, we compare against CRAFT~\cite{fel2023craft} and CONE-SHAP~\cite{li2021instance} while for object localization, we consider SLIC~\cite{SLIC}, LOST~\cite{LOST}, TokenCut~\cite{wang2022tokencut}, and the model by \cite{lv2024weakly} as baselines.
\textbf{Hardware.}
All experiments run on a dual AMD Epyc machine with 2 $\times$ 64 cores with 2.25~GHz, 2~TiB of memory, and NVIDIA A100 GPUs with 80 GB memory.
\looseness=-1

\subsection{Accuracy}\label{sec:acc}
To determine whether NAVE's concept representations contain discriminative information, their quality and informativeness are evaluated in a classification task.
\looseness=-1

A Random Forest (RF) classifier is trained to predict the labels based on the concept representation. The backbone model is a ResNet50 pretrained on ImageNet1K and subsequently fine-tuned on each dataset. Net2Vec, CRAFT, and CONE-SHAP are used as baselines. The concept representations for the baselines are computed from the output of the last block of the encoder, consistent with the information used by N4. For NAVE, N4, N34, and N234 are compared. For the baselines, the RF trained on the explanation maps, whereas for NAVE it uses the cluster proportions in the explanations. For all methods, one internal clustering is pretrained with $K=20$ on 500 training images per class (30 for CUB200\footnote{CUB200 has only 30 training instances per class.}). The class-clusterings are then merged before testing. For MNIST, this results in 10 $k$-means with $K=20$ merged into a single $k$-means with $ K=20\times10=200$. Accuracy scores in percentage are reported in Table~\ref{tab:acc} and an ablation study is provided in Appendix~C. %~\ref{apx-sec:abl}. 
Further, lines 9 and 10 of Algorithm~\ref{alg:nave} require normalization, which we justify experimentally, presented in Table~\ref{tab:norm+scal}.
Appendix~F includes more detailed results with object localization as well.

\begin{table}[t]
  \centering
  \caption{Test accuracy scores on MNIST, STL10, and CUB200 of a ResNet50 along with that of a Random Forest classifier trained on either the concept decomposition computed by Net2Vec, CRAFT and CONE-SHAPE, or, for three NAVE settings N234, N34 and N4, on the cluster proportions. Scores larger than that of the ResNet50 are marked in bold.}
  \label{tab:acc}
  {   \small \centering
          \begin{tabular*}{\linewidth}{@{\extracolsep{\fill}}lccccccc}
                  \toprule
Dataset     & \textit{ResNet50} & Net2Vec & CRAFT & CONE-SHAP & N234 & N34 & N4\\ \midrule
MNIST       & \textit{99.4} & 99.4 & 99.4 & 10.0 & 98.7 & 99.1 & 99.3 \\
STL10       & \textit{87.6} & \textbf{90.1} & \textbf{88.6} & 10.4 & 86.2 & \textbf{89.5} & \textbf{88.7} \\
CUB200      & \textit{56.8} & \textbf{58.5} & \textbf{57.2} &  1.1 & 54.6 & \textbf{66.0} & \textbf{61.6} \\
                  \bottomrule
          \end{tabular*}}
\end{table}

\begin{table}[t]
  \centering
  \caption{Ablation study of the normalization (N) and scaling (S) in the NAVE algorithm. Here we replicate the experiments of Table~\ref{tab:acc} with or without normalization or scaling.}
  \label{tab:norm+scal}
  {   \small \centering
          \begin{tabular*}{\linewidth}{@{\extracolsep{\fill}}lccccccccccc}
                  \toprule
  %\multicolumn{12}{c}{Classifcation} \\
Dataset     & \textit{ResNet50} & & \multicolumn{4}{c}{N234} & & \multicolumn{4}{c}{N34}\\
            &                   & & N+S & S & N & - & & N+S & S & N & - \\  \midrule
MNIST       & \textit{99.4}     & & 98.7 & 99.4 & 99.3 & 99.3 & & 99.1 & 99.2 & 99.3 & 99.2 \\
STL10       & \textit{87.6}     & & 86.2 & 88.7 & 90.2 & 88.2 & & 89.5 & 88.0 & 90.0 & 88.3 \\
                  \bottomrule
          \end{tabular*}}
\end{table}

\textbf{Discussion.} All methods except CONE-SHAP achieve accuracy comparable to or better than that of the backbone model, which is expected since they leverage the final encoder output also used by the ResNet classifier. 
The limited performance of CONE-SHAP could be attributed to the patch-projections being out of distribution, which in turn affects the quality of the concept assignments. On STL10 and CUB200, N34, and N4 outperform the backbone, while N234 does not. We conjecture that the shallower block brings less compressed information confusing the RF. At the same time, N34 performs better than N4. The optimal performance requires thus balancing different depth. Overall, most concept decompositions except CONE-SHAP carry sufficient information for classification.

\subsection{Object Localization}\label{sec:wsol}

Since the rationale is to group together parts of the image deemed similar, the explanation maps are expected to contain semantic information. A variant of the weakly supervised object localization (WSOL) task \cite{LOST} is proposed as a proxy to evaluate and validate the alignment between the extracted concepts and the semantic content of the input.

The original WSOL task evaluates the covering of bounding boxes around target object by quantized attribution maps. However, because NAVE is not an attribution method, we select instead the segment with the bounding box exhibiting the largest Intersection over Union (IoU) with the target bounding box. This approach overestimates the WSOL performance of NAVE and aims to evaluate how well it is able to isolate the semantics of the target object. As segments may not be convex, using the outer bounding box can introduce artifacts. Therefore, the largest inner box is computed. While this approach guarantees intersection with the ground truth box, the overlap may be reduced. Additional details and examples of these strategies are provided in Appendix~E. 
\looseness=-1

Explanation maps are computed on a per-image basis (batch size of one) using K=10 and the output of the last block for CRAFT and CONE-SHAP, and K=5 for NAVE N234, N34, and N4. The backbone architecture is a frozen ResNet50 pretrained on DINO. For comparison, SLIC is also included in the analysis. Following the protocol described in \cite{LOST} on the VOC07 and VOC12 datasets, Table~\ref{tab:wsol-r50} reports the Average Precision at 50\% (AP@50\%), also referred to as the Correct Localization metric. This metric represents the frequency with which predicted boxes achieve an Intersection over Union (IoU) greater than 50\% with at least one ground truth bounding box. \looseness=-1 \\

\begin{table}[t]
  \centering
  \caption{Object localization performance measured using AP@50\% on VOC07, VOC12 of a ResNet50 trained on DINO. The internal clusterings of CONE-SHAP and CRAFT produce $K=10$ clusters. For NAVE, $K=5$ and we compare three settings N234, N34 and N4. For reference, we include scores obtained using SLIC.}
  \label{tab:wsol-r50}
  {   \small \centering
          \begin{tabular*}{\linewidth}{@{\extracolsep{\fill}}lcccccc}
                  \toprule
Dataset     & SLIC & CRAFT & CONE-SHAP & N234 & N34 & N4\\ \midrule
VOC07       & 43.1 & 30.5 & 47.3 & 60.5 & 63.5 & 59.7 \\
VOC12       & 46.0 & 26.6 & 48.3 & 61.7 & 65.0 & 62.2 \\
                  \bottomrule
          \end{tabular*}}
\end{table}

\textbf{Discussion.} 
The three NAVE configurations outperform the baselines by more than 10 percentage points. Notably, CONE-SHAP explanations, which internally use SLIC segmentations, achieve higher scores than standard SLIC. The performance gap between NAVE and other methods, particularly CRAFT, contextualizes the results presented in Table~\ref{tab:acc}. Although labels can be derived from the concept decompositions of CRAFT, they do not align effectively with semantic information. These findings corroborate the previously observed (Figure~\ref{fig:expl}) poor quality of the explanation maps produced by CRAFT and CONE-SHAP.
\looseness=-1

Table~\ref{tab:wsol-vit} reports the same experiment on a ViT-S/16 trained on DINO, alongside state-of-the-art WSOL methods that rely on backpropagation to guide segmentation. 
Without any prediction signal (no labels, no gradients), NAVE nonetheless outperforms DINO-seg and LOST~\cite{LOST} on VOC07, VOC12, and COCO20k. This is a notable result: NAVE achieves competitive object localization as a side effect of encoder inspection, not as its primary objective. Methods such as TokenCut~\cite{wang2022tokencut} and WSCUOD~\cite{lv2024weakly} that leverage backpropagation to guide segmentation remain stronger, but they are not post-hoc inspection tools.
\looseness=-1

\begin{table}[t]
  \centering
  \caption{Object localization performance measured using AP@50\% on VOC07, VOC12, and COCO20k of a ViT-S/16 trained on DINO. The number of clusters is set to $K=10$ and $K=5$ for CONE-SHAPE and NAVE, respectively and both use the output of the last transformer block (N:12).
  Scores from cited works are taken from their works.
  }
  \label{tab:wsol-vit}
  {   \small \centering
          \begin{tabular*}{\linewidth}{@{\extracolsep{\fill}}lcccccc}
                  \toprule
Dataset     & \small{DINO-seg\cite{LOST}} & \small{LOST\cite{LOST}}  & \small{TokenCut\cite{wang2022tokencut}} & \small{WSCUOD\cite{lv2024weakly}} & \small{CONE-SHAP} & \small{NAVE} \\ \midrule
VOC07   & 45.8 & 61.9 & 68.8 & 70.6 & 51.6 & 63.1 \\
VOC12   & 46.2 & 64.0 & 72.1 & 72.1 & 49.6 & 64.3 \\
COCO20k & 42.1 & 50.7 & 58.8 & 63.5 & 52.4 & 61.0 \\
                  \bottomrule
          \end{tabular*}}
\end{table}

\begin{table}[t]
  \centering
  \caption{Unsupervised segmentation accuracy and mean intersection over union (mIoU) on a subset of COCO20k with a ResNet18 pretrained on ImageNet. 
  }
  \label{tab:useg}
  {   \small \centering
          \begin{tabular*}{\linewidth}{@{\extracolsep{\fill}}llccc|cccc}
                  \toprule
    & & \small{Mod. DC}\cite{cho2021picie} & \small{IIC}\cite{IIC}  & \small{PiCIE}\cite{cho2021picie} & \small{PiCIE-max} & \small{N234-max} & \small{N34-max} & \small{N4-max} \\ \midrule
Acc  &        & 32.2 & 21.8 & 50.3 & 69.9 & \textbf{93.5} & 81.0 & 76.3 \\
mIoU &        & 9.8  & 6.7  & 14.9 & 31.5 & \textbf{82.6} & 51.8 & 42.7 \\
                  \bottomrule
          \end{tabular*}}
\end{table}

\subsection{Unsupervised Segmentation}

While NAVE’s main goal is not unsupervised semantic segmentation, we evaluate its potential for this task. Since concept extraction is unsupervised and matching concepts to dataset classes is outside the scope of this study, segments are labeled with their most likely class per image. This yields an upper bound on the performance of NAVE-based unsupervised segmentation. Following \cite{cho2021picie}, Table~\ref{tab:useg} compares PiCIE, IIC~\cite{IIC}, and a modified DC~\cite{caron2018deep}, with label-segment maximizing matching also applied to PiCIE (PiCIE-max). The backbone is a ResNet18 pretrained on ImageNet1K~\cite{russakovsky2015imagenet}. We report results for N234-max, N34-max and N4-max with $K=20$ clusters fitted on each test batch. Results without label-matching maximization and with clustering fitted only on the first test batch are reported in Appendix~H.

The strongest results are observed for N234-max; however, several factors should be considered. Segment assignments are optimized for each test image to maximize accuracy. Also, N234 produces more complex explanations with smaller segments which boosts accuracy and mIoU after labeled assignments (see Figure 11 in Appendix~H). Nonetheless, the performance gap between PiCIE-max and N4-max suggests that NAVE’s explanation maps align better with ground-truth maps. With a dedicated concept-to-cluster assignment, NAVE-based unsupervised segmentation may have the potential to achieve competitive results.

\begin{table*}[t]
  \centering
  \caption{
  Localization performance (AP@50\%) of the various design and training choices.\looseness=-1}

  \label{tab:training}
  {   \small \centering
          \begin{tabular*}{\linewidth}{@{\extracolsep{\fill}}llccc}
                  \toprule
& Feature & VOC07 & VOC12 & COCO20k \\ \midrule
\multirow{6}{*}{\rotatebox[origin=c]{90}{\parbox{4em}{\centering a. Training \\ Dataset}}}
%\midrule
& ImageNet1K       & {\bf68.7} & {\bf69.5} & {\bf64.1} \\
& STL-10           & 44.2 & 44.7 & 51.3 \\ 
& Chest-X-Ray      & 42.8 & 45.1 & 42.7 \\ 
& Random Initialization & 45.2 & 46.9 & 48.0
\\[2mm] 
& DINO             & 62.7  & 62.3  & 61.0 \\ 
& SLIC             & 43.1 & 46.0 & 38.5 \\ 
                  \bottomrule
\multirow{4}{*}{\rotatebox[origin=c]{90}{\parbox{4em}{\centering b. Training \\ Scheme}}}
& Random Initialization        & 45.2  & 46.9  & 48.0 \\ 
& Supervised (ImageNet1K) & {\bf68.7}  & {\bf69.5}  & 64.1 \\ 
& SSL DINO            & 63.1  & 64.3  & 60.6 \\ 
& SSL DINOv2     & 66.0  & 66.2  & {\bf65.1} \\ 
                  \bottomrule
\multirow{3}{*}{\rotatebox[origin=c]{90}{\parbox{3em}{\centering c. Arch.}}}
& ResNet50 (ImageNet1K)  & 61.2  & 62.3  & - \\ 
& ViT-S/16 (ImageNet1K)   & {\bf68.7}  & {\bf69.5}  & {\bf64.1} \\ 
& ViT-B/16 (ImageNet1K)   & 44.1  & 45.3  & 48.6 \\ 
                  \bottomrule
          \end{tabular*}}
\end{table*}

\subsection{The Influence of Training}

This set of experiments examines how the training protocol affects the object localization performance of NAVE explanations. The number of clusters is fixed at $K=5$ for all experiments. The NAVE layer setting is N34 for ResNet50 and N:12 (last output) for the ViTs. \looseness=-1

\subsubsection{Training Set}\label{sec:exp-set}

The role of the training dataset in shaping the concepts extracted by the image encoder is investigated. The underlying hypothesis is that a model lacking exposure to specific concepts, such as a person, is unlikely to extract or recognize them, leading to their absence in NAVE outputs. This hypothesis is evaluated using the object localization protocol by comparing an untrained (randomly initialized) ViT-S/16-based classifier with three models trained on ImageNet1K, STL-10, or Chest-X-Ray. Results in Table~\ref{tab:training}.a indicate that training on ImageNet1K consistently produces the highest AP@50\% across all datasets, even surpassing a DINO encoder. Classifiers trained on smaller (STL-10) or out-of-domain (Chest-X-Ray) datasets achieve scores comparable to the untrained model. Notably, predictions based on SLIC segmentations yield performance equivalent to that of explanations from an untrained model. These results confirm that the quality of NAVE explanations is tied to the encoder's training.
\looseness=-1 

\subsubsection{Training Scheme}\label{sec:exp-scheme}

NAVE is used to demonstrate how different training schemes yield encoders with distinct behaviors. Table~\ref{tab:training}.b presents AP@50\% for a ViT-S/16 that is either untrained (random initialization), trained for classification (supervised ImageNet1K), or trained using self-supervised learning (SSL) with DINO and DINOv2 frameworks. The DINOv2 model utilizes a ViT-S/14 architecture.
Supervised and self-supervised learning approaches do not demonstrate a clear advantage over one another. Indeed, DINOv2 achieves performance comparable to fully supervised training. However, DINOv2 demands substantial computational resources and a significantly larger image dataset.
\looseness=-1

\subsubsection{Architecture}\label{sec:archi}
The influence of architecture on NAVE is assessed by comparing ViT-S, ViT-B, and ResNet50 models trained on ImageNet1K, with localization scores presented in Table~\ref{tab:training}.c. 
ViT-B demonstrates unexpectedly poor performance despite its larger size. It is hypothesized that ViT-B's overparametrization leads to spurious concepts or subdivides objects into smaller, less meaningful components, which NAVE is unable to extract effectively. In contrast, ViT-S outperforms ResNet50 despite having a similar parameter budget. The architectural design of ViT-S provides better inductive biases, enabling the model to learn more generalizable concepts and achieve higher evaluation scores.

\begin{table}[t]
  \centering
  \caption{Influence of the {registers} on the AP@50\% performance.}
  \label{tab:registers}
  {   \small \centering
          \begin{tabular*}{\linewidth}{@{\extracolsep{\fill}}lcccc}
                  \toprule
Architecture & Registers?         & VOC07 & VOC12 & COCO20k \\ \midrule
 DINOv2-ViT-S/14 &            & 66.0  & {\bf67.1}  & {\bf66.2} \\ 
 DINOv2-ViT-S/14 & \checkmark & {\bf66.5}  & 66.4  & 65.7 \\ 
 DINOv2-ViT-B/14 &            & 63.8  & 64.0 & 64.0 \\ 
 DINOv2-ViT-B/14 & \checkmark & 65.7  & 66.8 & 65.4 \\ 
                  \bottomrule
          \end{tabular*}}
\end{table}

\section{NAVE for Inspection}\label{sec:inspect}
This section presents two empirical findings enabled by NAVE as an inspection tool. First we show that attention norm artifacts in ViTs are background-specific and do not corrupt object-level concepts. Second, we provide the first direct visualization of shortcut saturation in encoder representations.
We show that a learned shortcut dominates the encoder's concept structure at test time while leaving non-shortcut content largely intact.
These findings demonstrate that NAVE can reveal phenomena inside the encoder that remain invisible to prediction-based XAI methods.

\begin{wrapfigure}[21]{r}{.45\linewidth}
    \vspace{-2em}
    \centering
    \includegraphics[width=.95\linewidth]{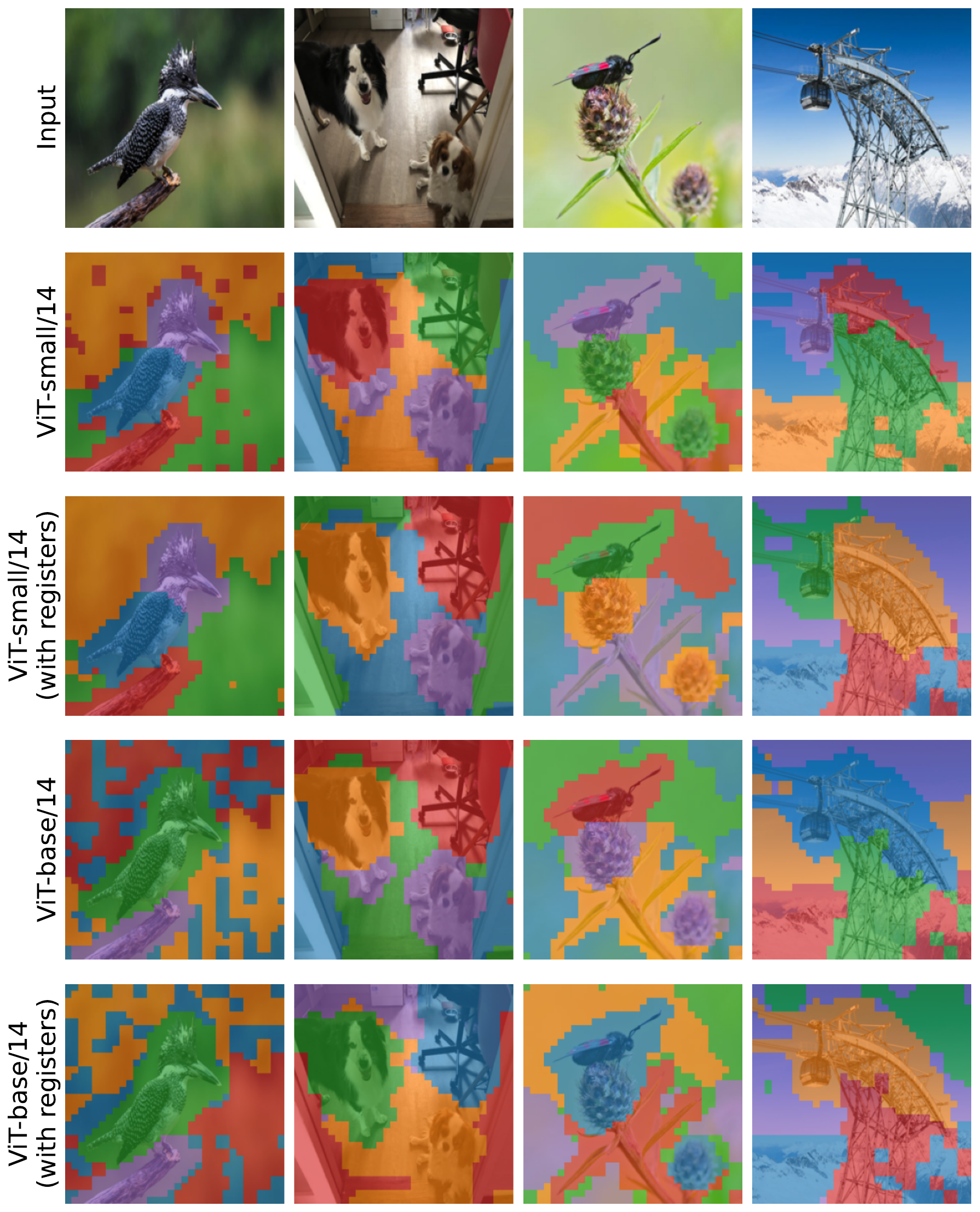}
    \caption{NAVE on four sample images (with $K=5$ clusters) reveals that registers do not fully safeguard ViTs trained with DINOv2 against artifacts.
    \looseness=-1
    }
    \label{fig:dinov2_registers}
\end{wrapfigure}
\subsection{ViT and Registers}\label{sec:vit-reg}

Recent studies on vision transformer training have identified challenges with learned representations in both supervised and self-supervised regimes. 
Darcet et al.~\cite{darcet2024vision} report that individual tokens may exhibit norms that differ substantially from others in the same layer, without correlation to the original inputs. NAVE characterizes this behavior as specific to the background.
\looseness=-1

Figure~\ref{fig:dinov2_registers} displays image-wise NAVE explanations for ViT-S/14 and ViT-B/14 ($K=5$, N:12), both with and without registers, using four examples take from the original study~\cite{darcet2024vision}. The main semantic content of the input images is consistently isolated by one or more concepts. For example, the flower heads in the third column are isolated by all four models. The first column shows that NAVE captures artifacts caused by exploding norms in the attention maps, which registers are designed to mitigate. These artifacts are either linked to a semantic concept (ViT-S, first image, red cluster) or form a separate cluster. These artifacts primarily affect the background.
Results in Table~\ref{tab:registers} for the four models indicate that adding registers does not significantly change AP@50\% localization scores.
This confirms that attention norm artifacts are background-specific and do not corrupt object-level concepts. The semantic integrity of foreground representations is thus more robust to training instabilities than previously assumed.
\looseness=-1

\subsection{Augmentation and Shortcuts}\label{sec:clever}

\begin{figure*}[!ht]
    \centering
    \includegraphics[width=\linewidth]{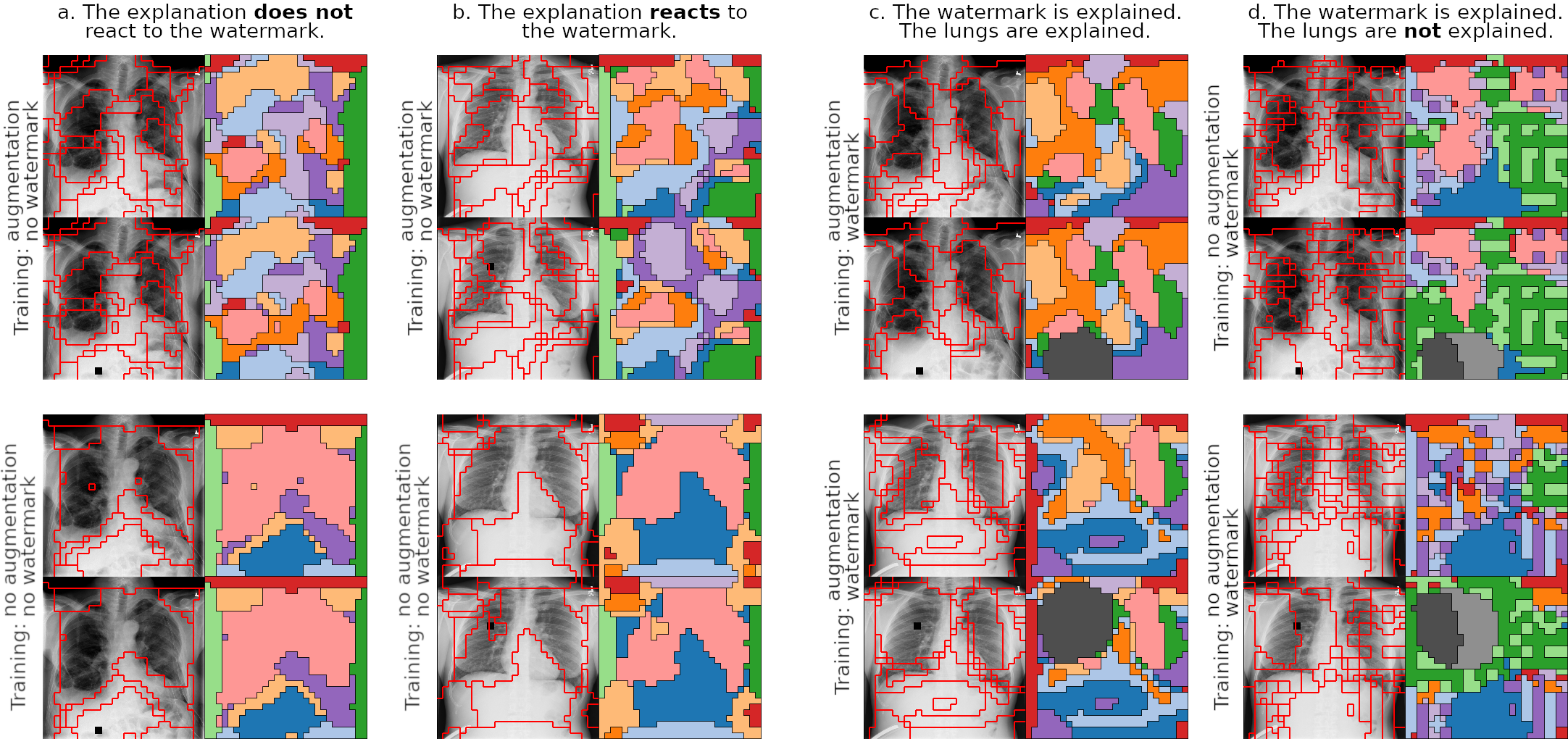}
    \caption{Explanations of a ResNet50 trained under various regimes of data augmentation and watermarking.
    % If concepts are capturing the watermark, they are disproportionately large compared to the watermark itself.
    When the watermark is present in training, a single concept saturates a neighborhood far larger than the watermark itself. This is direct evidence of shortcut saturation in the encoder. Notably, non-shortcut content (e.g., the lungs) remains largely intact in the remaining concepts.}
    \label{fig:pneumo}
\end{figure*}

This section employs NAVE to visualize the effects of augmentations and shortcuts~\cite{geirhos2020shortcut} using the Chest-X-Ray dataset. Following the methodology of \cite{gautam2022demonstrating}, a ResNet50 is trained for pneumonia prediction using data from a single hospital, and a modified dataset is generated in which all negative images are watermarked with a small black box at random locations. Four training strategies are evaluated: with or without augmentations and with or without watermarks. Two negative images are selected to illustrate the four model behaviors in Figure~\ref{fig:pneumo}. NAVE explanations are computed using the last two residual blocks (N34), with the clustering pretrained over 200 images using $K=10$.
\looseness=-1

For Figures~\ref{fig:pneumo}.a and b, models are trained on the original dataset (no watermarks), both with and without augmentations. Augmentations result in more complex and semantically organized explanations, with the lungs distinctly separated into upper and lower regions. Introducing a watermark to the left image does not alter the NAVE explanation, whereas it does for the second image, particularly when augmentations are used. This suggests the encoder's sensitivity to the watermark depends on what it obfuscates. For Figures~\ref{fig:pneumo}.c and d, models are trained on the watermarked dataset, again with and without augmentations. When a black box is present, both encoders exhibit strong responses with clusters associated to the watermark saturating a large neighborhood. Interestingly, the processing of the remaining image content is not much affected by the shortcut. These degenerate explanation maps highlight the risks associated with shortcuts in the training set and underscore the consequences of inadequate training strategies.
\looseness=-1

In summary, the complexity and quality of NAVE explanations reflect the robustness of the training scheme. When a shortcut is learned during training, its presence in a test image saturates NAVE’s explanation maps. Together, these use cases demonstrate how NAVE can be turned into a practical diagnostic tool, capable of surfacing both architectural artifacts and dataset-level biases directly from the encoder's internal representations.\looseness=-1

\section{Conclusion}

We introduced Neuro-Activated Vision Explanations (NAVE), a method for representing the concepts captured by the encoder of vision models. We demonstrated, both qualitatively and quantitatively, that NAVE explanation maps align with semantic content and can be leveraged for downstream tasks such as classification and model inspection. We presented two use cases for model inspection: first, we showed that artifacts resulting from exploding norms in attention maps of ViTs primarily affect image backgrounds; second, we visualized, for the first time, information saturation on explanation maps caused by shortcut connections. The main limitations are due to $k$-means, which can be mitigated using the mini-batch variant or replaced by hierarchical clustering at higher computational cost.

\bibliographystyle{splncs04}
\bibliography{references}

@article{russakovsky2015imagenet,
  title={Imagenet large scale visual recognition challenge},
  author={Russakovsky, Olga and Deng, Jia and Su, Hao and Krause, Jonathan and Satheesh, Sanjeev and Ma, Sean and Huang, Zhiheng and Karpathy, Andrej and Khosla, Aditya and Bernstein, Michael and others},
  journal={International Journal of Computer Vision},
  volume={115},
  pages={211--252},
  year={2015},
  publisher={Springer}
}

@inproceedings{darcet2024vision,
  title={Vision Transformers Need Registers},
  author={Darcet, Timoth{\'e}e and Oquab, Maxime and Mairal, Julien and Bojanowski, Piotr},
  booktitle={International Conference on Learning Representations},
  year={2024}
}

@article{jacob,
	title={From clustering to cluster explanations via neural networks},
	author={Kauffmann, Jacob and Esders, Malte and Ruff, Lukas and Montavon, Gr{\'e}goire and Samek, Wojciech and M{\"u}ller, Klaus-Robert},
	journal={IEEE Transactions on Neural Networks and Learning Systems},
	year={2022},
	publisher={IEEE}
}

@inproceedings{ViT,
    title={An Image is Worth 16x16 Words: Transformers for Image Recognition at Scale},
    author={Alexey Dosovitskiy and Lucas Beyer and Alexander Kolesnikov and Dirk Weissenborn and Xiaohua Zhai and Thomas Unterthiner and Mostafa Dehghani and Matthias Minderer and Georg Heigold and Sylvain Gelly and Jakob Uszkoreit and Neil Houlsby},
    booktitle={International Conference on Learning Representations},
    year={2021},
}

@inproceedings{DINO,
	title={Emerging properties in self-supervised vision transformers},
	author={Caron, Mathilde and Touvron, Hugo and Misra, Ishan and J{\'e}gou, Herv{\'e} and Mairal, Julien and Bojanowski, Piotr and Joulin, Armand},
	booktitle={Proceedings of the IEEE/CVF International Conference on Computer Vision},
	pages={9650--9660},
	year={2021}
}

@article{DINOv2,
title={{DINO}v2: Learning Robust Visual Features without Supervision},
author={Maxime Oquab and Timoth{\'e}e Darcet and Th{\'e}o Moutakanni and Huy V. Vo and Marc Szafraniec and Vasil Khalidov and Pierre Fernandez and Daniel HAZIZA and Francisco Massa and Alaaeldin El-Nouby and Mido Assran and Nicolas Ballas and Wojciech Galuba and Russell Howes and Po-Yao Huang and Shang-Wen Li and Ishan Misra and Michael Rabbat and Vasu Sharma and Gabriel Synnaeve and Hu Xu and Herve Jegou and Julien Mairal and Patrick Labatut and Armand Joulin and Piotr Bojanowski},
journal={Transactions on Machine Learning Research},
issn={2835-8856},
year={2024},
}

@inproceedings{LOST,
   title = {Localizing Objects with Self-Supervised Transformers and no Labels},
   author = {Oriane Sim\'eoni and Gilles Puy and Huy V. Vo and Simon Roburin and Spyros Gidaris and Andrei Bursuc and Patrick P\'erez and Renaud Marlet and Jean Ponce},
   booktitle = {Proceedings of the British Machine Vision Conference},
   month = {November},
   year = {2021}
}

@inproceedings{wang2022tokencut,
      title={Self-supervised Transformers for Unsupervised Object Discovery using Normalized Cut},
      author={Wang, Yangtao and Shen, Xi and Hu, Shell Xu and Yuan, Yuan and Crowley, James L. and Vaufreydaz, Dominique},
      booktitle={Conference on Computer Vision and Pattern Recognition},
      year={2022}
    }

@article{lv2024weakly,
  title={Weakly-supervised contrastive learning for unsupervised object discovery},
  author={Lv, Yunqiu and Zhang, Jing and Barnes, Nick and Dai, Yuchao},
  journal={IEEE Transactions on Image Processing},
  year={2024},
  publisher={IEEE}
}

@inproceedings{zeiler2014visualizing,
	title={Visualizing and understanding convolutional networks},
	author={Zeiler, Matthew D and Fergus, Rob},
	booktitle={European Conference on Computer Vision},
	pages={818--833},
	year={2014},
	organization={Springer}
}

@article{LRP,
	title={On pixel-wise explanations for non-linear classifier decisions by layer-wise relevance propagation},
	author={Bach, Sebastian and Binder, Alexander and Montavon, Gr{\'e}goire and Klauschen, Frederick and M{\"u}ller, Klaus-Robert and Samek, Wojciech},
	journal={PloS ONE},
	volume={10},
	number={7},
	pages={e0130140},
	year={2015},
	publisher={Public Library of Science San Francisco, CA USA}
}

@inproceedings{IG,
	title={Axiomatic attribution for deep networks},
	author={Sundararajan, Mukund and Taly, Ankur and Yan, Qiqi},
	booktitle={International Conference on Machine Learning},
	pages={3319--3328},
	year={2017},
	organization={PMLR}
}

@inproceedings{GC,
	title={Grad-{CAM}: Visual explanations from deep networks via gradient-based localization},
	author={Selvaraju, Ramprasaath R and Cogswell, Michael and Das, Abhishek and Vedantam, Ramakrishna and Parikh, Devi and Batra, Dhruv},
	booktitle={Proceedings of the IEEE International Conference on Computer Vision},
	pages={618--626},
	year={2017}
}

@article{SLIC,
	title={{SLIC} superpixels compared to state-of-the-art superpixel methods},
	author={Achanta, Radhakrishna and Shaji, Appu and Smith, Kevin and Lucchi, Aurelien and Fua, Pascal and S{\"u}sstrunk, Sabine},
	journal={IEEE Transactions on Pattern Analysis and Machine Intelligence},
	volume={34},
	number={11},
	pages={2274--2282},
	year={2012},
	publisher={IEEE}
}

@inproceedings{cho2021picie,
	title={{PiCIE}: Unsupervised semantic segmentation using invariance and equivariance in clustering},
	author={Cho, Jang Hyun and Mall, Utkarsh and Bala, Kavita and Hariharan, Bharath},
	booktitle={Proceedings of the IEEE/CVF Conference on Computer Vision and Pattern Recognition},
	pages={16794--16804},
	year={2021}
}

@techreport{erhan2009visualizing,
	title={Visualizing higher-layer features of a deep network},
	author={Erhan, Dumitru and Bengio, Yoshua and Courville, Aaron and Vincent, Pascal},
	institution={University of Montreal},
	year={2009}
}

@article{DeepDream,
	title={Deep inside convolutional networks: Visualising image classification models and saliency maps},
	author={Simonyan, Karen and Vedaldi, Andrea and Zisserman, Andrew},
	journal={arXiv preprint arXiv:1312.6034},
	year={2013}
}

@inproceedings{ResNet,
	title={Deep residual learning for image recognition},
	author={He, Kaiming and Zhang, Xiangyu and Ren, Shaoqing and Sun, Jian},
	booktitle={Proceedings of the IEEE Conference on Computer Vision and Pattern Recognition},
	pages={770--778},
	year={2016}
}

@inproceedings{pytorch,
	title={{PyTorch}: An imperative style, high-performance deep learning library},
	author={Paszke, Adam and Gross, Sam and Massa, Francisco and Lerer, Adam and Bradbury, James and Chanan, Gregory and Killeen, Trevor and Lin, Zeming and Gimelshein, Natalia and Antiga, Luca and others},
	booktitle={Advances in Neural Information Processing Systems},
	volume={32},
	year={2019}
}

@inproceedings{coates2011analysis,
	title={An analysis of single-layer networks in unsupervised feature learning},
	author={Coates, Adam and Ng, Andrew and Lee, Honglak},
	booktitle={Proceedings of the 14th International Conference on Artificial Intelligence and Statistics},
	pages={215--223},
	year={2011}
}

@article{borys2023explainable,
  title={Explainable AI in medical imaging: An overview for clinical practitioners--Beyond saliency-based XAI approaches},
  author={Borys, Katarzyna and Schmitt, Yasmin Alyssa and Nauta, Meike and Seifert, Christin and Kr{\"a}mer, Nicole and Friedrich, Christoph M and Nensa, Felix},
  journal={European Journal of Radiology},
  volume={162},
  pages={110786},
  year={2023},
  publisher={Elsevier}
}

@inproceedings{williford2020explainable,
  title={Explainable face recognition},
  author={Williford, Jonathan R and May, Brandon B and Byrne, Jeffrey},
  booktitle={European Conference on Computer Vision},
  pages={248--263},
  year={2020},
  organization={Springer}
}

@article{panfilova2024applying,
  title={Applying explainable artificial intelligence methods to models for diagnosing personal traits and cognitive abilities by social network data},
  author={Panfilova, Anastasia S and Turdakov, Denis Yu},
  journal={Scientific Reports},
  volume={14},
  number={1},
  pages={5369},
  year={2024},
  publisher={Nature Publishing Group UK London}
}

@inproceedings{fel2023craft,
  title={Craft: Concept recursive activation factorization for explainability},
  author={Fel, Thomas and Picard, Agustin and Bethune, Louis and Boissin, Thibaut and Vigouroux, David and Colin, Julien and Cad{\`e}ne, R{\'e}mi and Serre, Thomas},
  booktitle={Proceedings of the IEEE/CVF Conference on Computer Vision and Pattern Recognition},
  pages={2711--2721},
  year={2023}
}

@inproceedings{kim2018interpretability,
  title={Interpretability beyond feature attribution: Quantitative testing with concept activation vectors (tcav)},
  author={Kim, Been and Wattenberg, Martin and Gilmer, Justin and Cai, Carrie and Wexler, James and Viegas, Fernanda and Sayres, Roryand},
  booktitle={International Conference on Machine Learning},
  pages={2668--2677},
  year={2018},
  organization={PMLR}
}

@inproceedings{SAM,
  title={Segment anything},
  author={Kirillov, Alexander and Mintun, Eric and Ravi, Nikhila and Mao, Hanzi and Rolland, Chloe and Gustafson, Laura and Xiao, Tete and Whitehead, Spencer and Berg, Alexander C and Lo, Wan-Yen and Doll{\'a}r, Piotr and Girshick, Ross},
  booktitle={Proceedings of the IEEE/CVF International Conference on Computer Vision},
  pages={4015--4026},
  year={2023}
}

@misc{pascal-voc-2007,
author = "Everingham, M. and Van~Gool, L. and Williams, C. K. I. and Winn, J. and Zisserman, A.",
title = "The {PASCAL} {V}isual {O}bject {C}lasses {C}hallenge 2007 {(VOC2007)} {R}esults",
year={2007}}

@misc{pascal-voc-2012,
author = "Everingham, M. and Van~Gool, L. and Williams, C. K. I. and Winn, J. and Zisserman, A.",
title = "The {PASCAL} {V}isual {O}bject {C}lasses {C}hallenge 2012 {(VOC2012)} {R}esults",
year={2012}}

@inproceedings{lin2014microsoft,
  title={Microsoft coco: Common objects in context},
  author={Lin, Tsung-Yi and Maire, Michael and Belongie, Serge and Hays, James and Perona, Pietro and Ramanan, Deva and Doll{\'a}r, Piotr and Zitnick, C Lawrence},
  booktitle={European Conference on Computer Vision},
  pages={740--755},
  year={2014},
  organization={Springer}
}

@article{gordon1987review,
  title={A review of hierarchical classification},
  author={Gordon, Allan D},
  journal={Journal of the Royal Statistical Society: Series A (General)},
  volume={150},
  number={2},
  pages={119--137},
  year={1987},
  publisher={Wiley Online Library}
}

@article{arias2024analysis,
  title={Analysis of the Clever Hans effect in COVID-19 detection using Chest X-Ray images and {B}ayesian Deep Learning},
  author={Arias-Londo{\~n}o, Juli{\'a}n D and Godino-Llorente, Juan I},
  journal={Biomedical Signal Processing and Control},
  volume={90},
  pages={105831},
  year={2024},
  publisher={Elsevier}
}

@article{geirhos2020shortcut,
  title={Shortcut learning in deep neural networks},
  author={Geirhos, Robert and Jacobsen, J{\"o}rn-Henrik and Michaelis, Claudio and Zemel, Richard and Brendel, Wieland and Bethge, Matthias and Wichmann, Felix A},
  journal={Nature Machine Intelligence},
  volume={2},
  number={11},
  pages={665--673},
  year={2020},
  publisher={Nature Publishing Group UK London}
}

@inproceedings{gautam2022demonstrating,
  title={Demonstrating the risk of imbalanced datasets in chest x-ray image-based diagnostics by prototypical relevance propagation},
  author={Gautam, Srishti and H{\"o}hne, Marina M-C and Hansen, Stine and Jenssen, Robert and Kampffmeyer, Michael},
  booktitle={2022 IEEE 19th International Symposium on Biomedical Imaging (ISBI)},
  pages={1--5},
  year={2022},
  organization={IEEE}
}

@article{kmex,
  title={Prototypical Self-Explainable Models Without Re-training},
  author={Gautam, Srishti and Boubekki, Ahcene and H{\"o}hne, Marina MC and Kampffmeyer, Michael},
  journal={Transactions on Machine Learning Research},
  year={2024},
}

@inproceedings{alvarez2018towards,
  title={Towards robust interpretability with self-explaining neural networks},
  author={Alvarez Melis, David and Jaakkola, Tommi},
  booktitle={Advances in Neural Information Processing Systems},
  volume={31},
  year={2018}
}

@inproceedings{protopnet,
  title={This looks like that: deep learning for interpretable image recognition},
  author={Chen, Chaofan and Li, Oscar and Tao, Daniel and Barnett, Alina and Rudin, Cynthia and Su, Jonathan K},
  booktitle={Advances in Neural Information Processing Systems},
  volume={32},
  year={2019}
}

@article{chormai2024disentangled,
  title={Disentangled explanations of neural network predictions by finding relevant subspaces},
  author={Chormai, Pattarawat and Herrmann, Jan and M{\"u}ller, Klaus-Robert and Montavon, Gr{\'e}goire},
  journal={IEEE Transactions on Pattern Analysis and Machine Intelligence},
  year={2024},
  publisher={IEEE}
}

@inproceedings{sun2023explain,
  title={Explain any concept: Segment anything meets concept-based explanation},
  author={Sun, Ao and Ma, Pingchuan and Yuan, Yuanyuan and Wang, Shuai},
  booktitle={Advances in Neural Information Processing Systems},
  volume={36},
  pages={21826--21840},
  year={2023}
}

@inproceedings{li2021instance,
  title={Instance-wise or class-wise? a tale of neighbor shapley for concept-based explanation},
  author={Li, Jiahui and Kuang, Kun and Li, Lin and Chen, Long and Zhang, Songyang and Shao, Jian and Xiao, Jun},
  booktitle={Proceedings of the 29th ACM International Conference on Multimedia},
  pages={3664--3672},
  year={2021}
}

@inproceedings{VCC,
  title={Visual concept connectome (vcc): Open world concept discovery and their interlayer connections in deep models},
  author={Kowal, Matthew and Wildes, Richard P and Derpanis, Konstantinos G},
  booktitle={Proceedings of the IEEE/CVF Conference on Computer Vision and Pattern Recognition},
  pages={10895--10905},
  year={2024}
}

@inproceedings{minikm,
  title={Web-scale k-means clustering},
  author={Sculley, David},
  booktitle={Proceedings of the 19th International Conference on World Wide Web},
  pages={1177--1178},
  year={2010}
}

@inproceedings{net2vec,
  title={Net2vec: Quantifying and explaining how concepts are encoded by filters in deep neural networks},
  author={Fong, Ruth and Vedaldi, Andrea},
  booktitle={Proceedings of the IEEE Conference on Computer Vision and Pattern Recognition},
  pages={8730--8738},
  year={2018}
}

@article{RabbitHull,
  title={Into the rabbit hull: From task-relevant concepts in DINO to minkowski geometry},
  author={Fel, Thomas and Wang, Binxu and Lepori, Michael A and Kowal, Matthew and Lee, Andrew and Balestriero, Randall and Joseph, Sonia and Lubana, Ekdeep S and Konkle, Talia and Ba, Demba and others},
  journal={arXiv preprint arXiv:2510.08638},
  year={2025}
}

@inproceedings{celeba,
  title = {Deep Learning Face Attributes in the Wild},
  author = {Liu, Ziwei and Luo, Ping and Wang, Xiaogang and Tang, Xiaoou},
  booktitle = {Proceedings of International Conference on Computer Vision (ICCV)},
  year = {2015} 
}

@inproceedings{hariharan2015hypercolumns,
  title={Hypercolumns for object segmentation and fine-grained localization},
  author={Hariharan, Bharath and Arbel{\'a}ez, Pablo and Girshick, Ross and Malik, Jitendra},
  booktitle={Proceedings of the IEEE Conference on Computer Vision and Pattern Recognition},
  pages={447--456},
  year={2015}
}

@inproceedings{IIC,
  title={Invariant information clustering for unsupervised image classification and segmentation},
  author={Ji, Xu and Henriques, Joao F and Vedaldi, Andrea},
  booktitle={Proceedings of the IEEE/CVF international conference on computer vision},
  pages={9865--9874},
  year={2019}
}

@inproceedings{caron2018deep,
  title={Deep clustering for unsupervised learning of visual features},
  author={Caron, Mathilde and Bojanowski, Piotr and Joulin, Armand and Douze, Matthijs},
  booktitle={Proceedings of the European conference on computer vision (ECCV)},
  pages={132--149},
  year={2018}
}

\begin{appendix}
	
	\section{Contents of the Supplementary Material}

	This supplementary material complements the main paper as follows.
	In Section~\ref{apx-sec:pca-km}, we provide additional examples comparing DINO/PCA and NAVE/$k$-means explanations as the number of components or clusters $K$ increases, on examples from CelebA, VOC07, CUB-200, and STL-10.
	In Section~\ref{apx-sec:acc}, we extend Table~2 with the classification accuracy obtained for different combinations of the number of clusters and the selected layers, including the CRAFT, CONE-SHAP, and Net2Vec baselines.
	In Section~\ref{apx-sec:normscale}, we report an ablation study of the normalization and scaling steps of Algorithm~1, replicating the experiments of Tables~2 and~3 with each component enabled or disabled.
	In Section~\ref{apx-sec:innerouterbox}, we detail the inner- and outer-box selection strategies used to derive a bounding box from a (possibly non-convex) NAVE segment, and illustrate their difference on a VOC07 image.
	In Section~\ref{apx-sec:wsol}, we extend Tables~5 and~6 with the object localization scores obtained using the outer-box strategy alongside the inner-box scores reported in the main text.
	In Section~\ref{apx-sec:wsolK}, we further extend Tables~5 and~6 with object localization scores for different numbers of clusters ($K=3$, $5$, and $7$), confirming that $K=5$ yields the best overall performance.
	In Section~\ref{apx-sec:useg}, we further extend Tables~5 with unsupervised segmentation scores for different numbers of clusters ($K=5$, $10$, $20$, and $27$) and different layer combinations.
	Finally, in Section~\ref{apx-sec:maps}, we present qualitative ablation studies of the explanation maps with respect to the number of clusters, the selected layers, and the clustering algorithm.

	\section{Comparison PCA and k-means}\label{apx-sec:pca-km}
	
	Additional examples comparing DINO/PCA and NAVE/$k$-means explanations as the number of components or clusters ($K$) increases. Both methods are applied to the output of the final transformer block (N:12) of a ViT-S/14 model trained on DINOv2. The examples are taken, respectively, from the CelebA, VOC07, CUB-200 and STL-10 datasets. In the PCA segmentations, pixels are assigned to the component with the greatest absolute value. The same color scheme is maintained across plots for the PCA components, with the first three directions in red, blue, and green, respectively. \looseness=-1

	\begin{figure}[!ht]
		\centering
		\includegraphics[width=\linewidth]{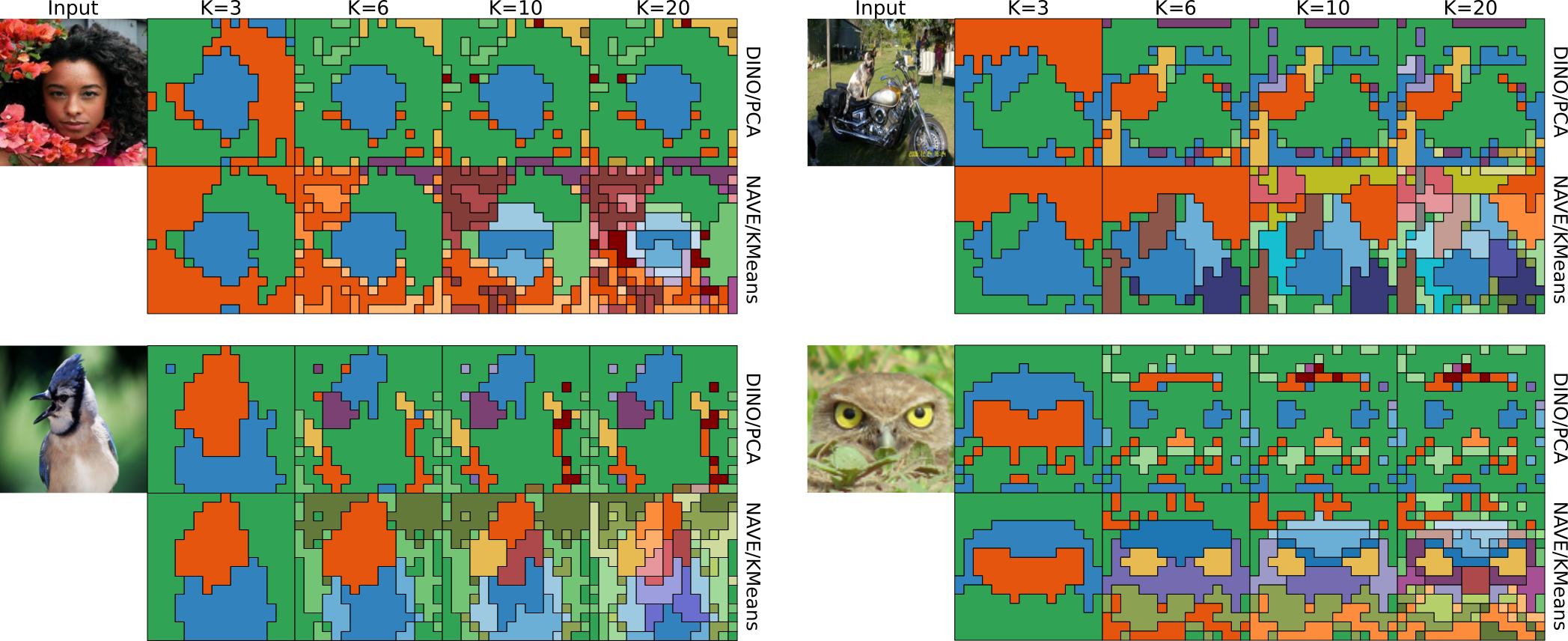}
		\caption{Explanations derived from DINO/PCA and NAVE/$k$-means with increasing number of components/clusters ($K$).
			\looseness=-1
		}
		\label{fig:pca_km_K}
	\end{figure}

	\section{Classification Accuracy}\label{apx-sec:acc}
	
\begin{figure}[!ht]
		\centering
		\includegraphics[width=.9\linewidth]{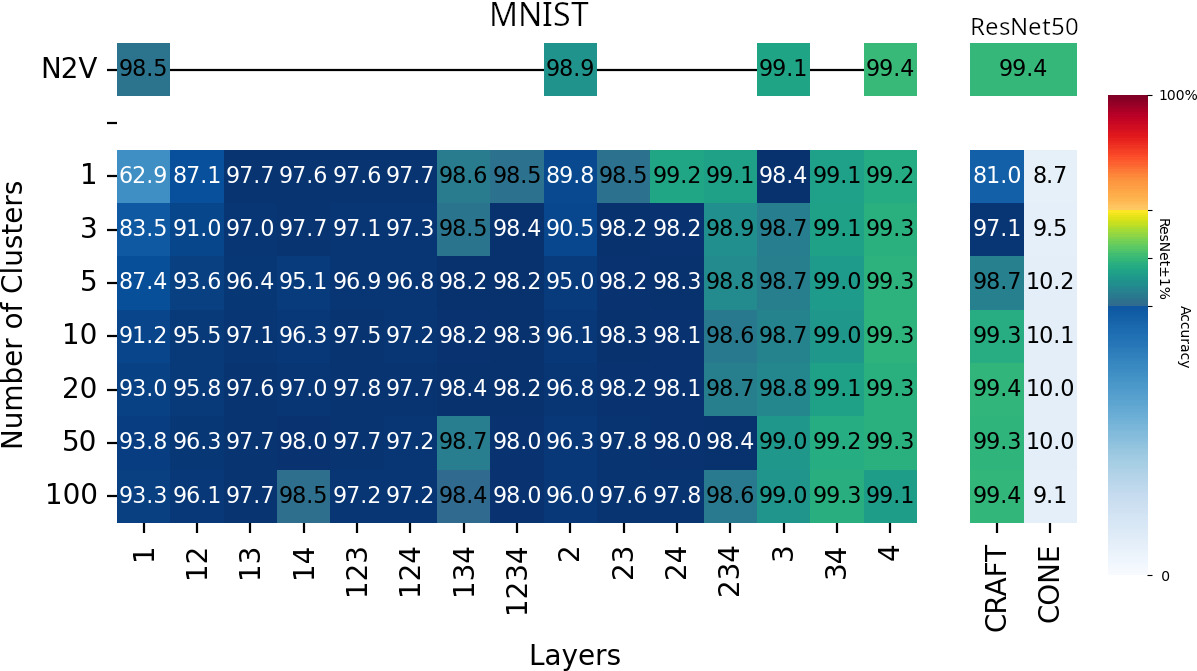} \\
		\label{apx-fig:acc}
	\end{figure}

	In Figure~\ref{apx-fig:acc}, we extend Table 2 of the main paper for different combinations of the number of clusters and the selected layers. We also include results for the baselines CRAFT, CONE-SHAP and Net2Vec. The accuracy achieved by the backbone ResNet50 is given in the upper right box.

	\begin{figure}[!ht]
		\centering
		\includegraphics[width=.9\linewidth]{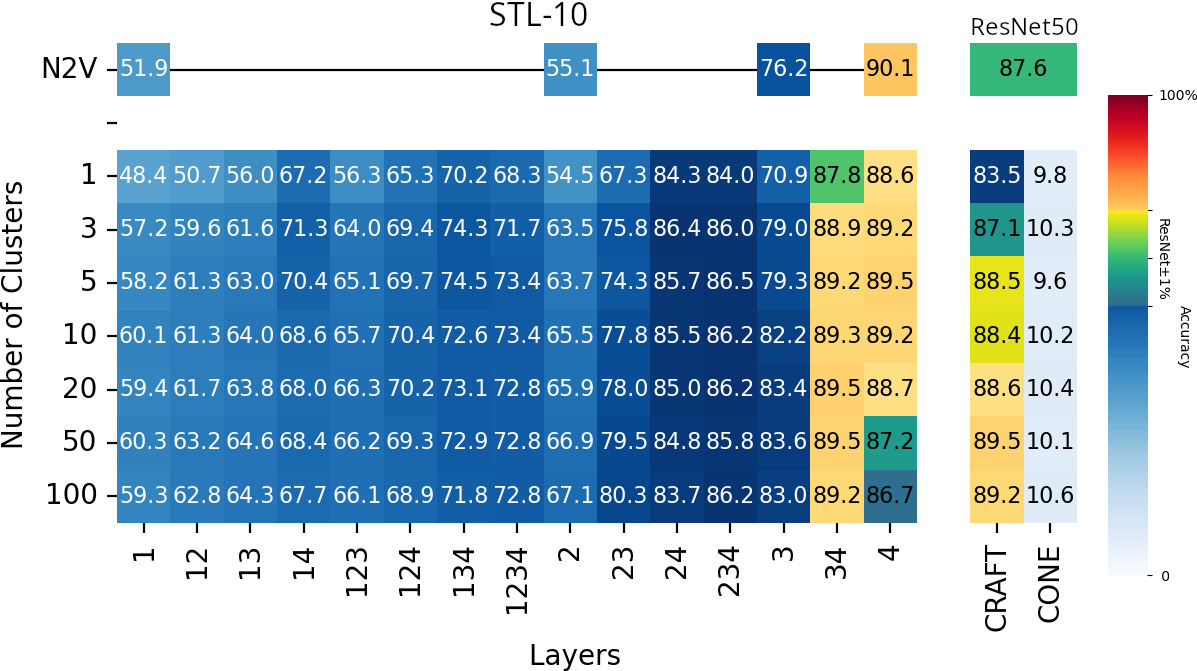} \\
		\includegraphics[width=.9\linewidth]{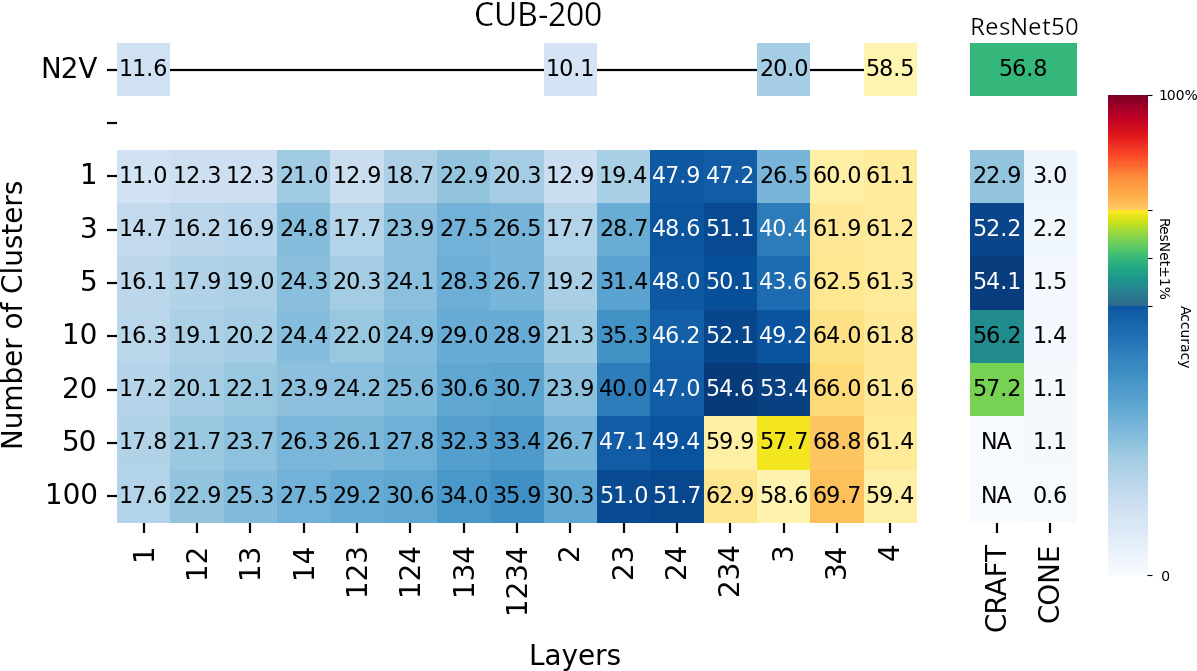} \\
		\caption{Test accuracy scores on MNIST, STL10, and CUB200 of a Random Forest classifier trained on either the concept decomposition computed by Net2Vec, CRAFT and CONE-SHAP, or, for multiple NAVE settings, on the cluster proportions. Scores lower, about equal and larger than that of the ResNet50 are colored in blue, green and orange/red, respectively.
		}
		\label{apx-fig:acc2}
	\end{figure}
	
	\newpage

	\section{Normalization and Scaling}\label{apx-sec:normscale}
	
	The normalization and the scaling occurring in lines 9 and 10, respectively of Algorithm 1 allow to balance the importance of the feature activations despite their different number of channels. Table~\ref{apx-tab:norm+scal} replicates the experiments of Tables~2 and 3 with or without the normalization (N) and scaling (S). The impact is milder on N34 because block 3 and 4 of the dilated ResNet50 return outputs with the same number of channels.
	
	\begin{table}[!ht]
		\centering
		\caption{Ablation study of the normalization (N) and scaling (S) in the NAVE algorithm. The upper and lower sections of the table replicate, respectively, the experiments of Tables 2 and 3 with or without normalization or scaling.}
		\label{apx-tab:norm+scal}
		{   \small \centering
			\begin{tabular*}{\linewidth}{@{\extracolsep{\fill}}lccccccccccc}
				\toprule
				\multicolumn{12}{c}{Classification} \\
				Dataset     & \textit{ResNet50} & & \multicolumn{4}{c}{N234} & & \multicolumn{4}{c}{N34}\\
				&                   & & N+S & S & N & - & & N+S & S & N & - \\  \midrule
				MNIST       & \textit{99.4}     & & 98.7 & 99.4 & 99.3 & 99.3 & & 99.1 & 99.2 & 99.3 & 99.2 \\
				STL10       & \textit{87.6}     & & 86.2 & 88.7 & 90.2 & 88.2 & & 89.5 & 88.0 & 90.0 & 88.3 \\ \midrule
				%CUB200      & \textit{56.8} & \textbf{58.5} & \textbf{57.2} &  1.1 & 54.6 & \textbf{66.0} & \textbf{61.6} \\
				\multicolumn{12}{c}{Object Localization} \\
				VOC07       &  -               & & 60.5 & 59.7 & 59.3 & 59.7 & & 63.5 & 63.5 & 63.4 & 63.5 \\
				\bottomrule
		\end{tabular*}}
	\end{table}

	\section{Inner and Outer Box Selection}\label{apx-sec:innerouterbox}
	
\begin{figure}[!ht]
		\centering
		\begin{tabular}{ccc}
			Original Image & Outer-box: & Inner-box: \\
			and ground truth & %\\
			IoU=$60.5\%$ & %\\
			IoU=$4.2\%$ \\
			\includegraphics[width=.3\linewidth]{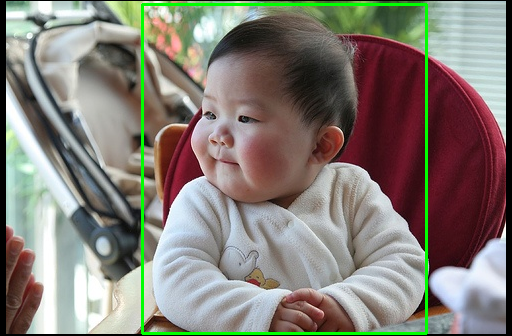} & %\\
			\includegraphics[width=.3\linewidth]{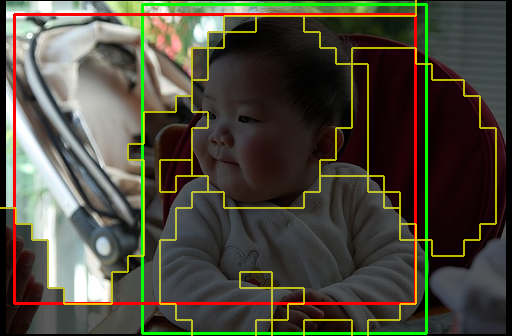} & %\\
			\includegraphics[width=.3\linewidth]{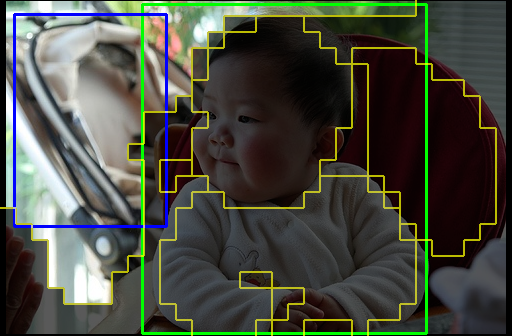} \\
		\end{tabular}
		%\\
		\caption{Example taken from VOC07 of the computed outer (red) and inner (blue) boxes of the top left segment along with IoU scores against the ground truth (green).
		}
		\label{fig:apx-boxes}
	\end{figure}

	The connected components of NAVE's segmentation might not be convex, hence using the bounding box (here called outer-box, Algorithm~\ref{alg:apx-outerbox}) might misguide what the component actually captures.
	Hence, we propose Algorithm~\ref{alg:apx-innerbox} which centers the box on the median of the area covered by the component.
	The inner-boxes are smaller and closer to what their component captures, yielding smaller IoU scores.
	We illustrate this phenomenon in Figure~\ref{fig:apx-boxes} on an image taken from VOC07.

	\begin{algorithm}[!ht]
		\caption{Python pseudocode to extract the outer-box}
		\label{alg:apx-outerbox}
		\begin{algorithmic}[1]
			\REQUIRE \texttt{Binary mask of dim. (H,W) }
			% \texttt{
				\STATE \texttt{coord = (mask>0).nonzero()}
				\STATE \texttt{xmin,xmax = min(coord[1]),max(coord[1])}
				\STATE \texttt{ymin,ymax = min(coord[0]),max(coord[0])}\\
				\RETURN{\texttt xmin,ymin,xmax,ymax}
				% }
		\end{algorithmic}
	\end{algorithm}

	\begin{algorithm}[!ht]
		\caption{Python pseudocode to extract the inner-box}
		\label{alg:apx-innerbox}
		\begin{algorithmic}[1]    
			\REQUIRE \texttt{Input mask of dim. (H,W)}
			\STATE \texttt{coord = (mask>0).nonzero()}
			\STATE \texttt{xs, ys = median(coord[1]), median(coord[0])}
			\STATE \texttt{xd = min(abs(xs-min(coord[1]), abs(xs-max(coord[1])))}
			\STATE \texttt{yd = min(abs(ys-min(coord[0])), abs(ys-max(coord[0])))}
			\STATE \texttt{xmin, xmax = clip(xs-xd,0,W), clip(xs+xd,0,W)}
			\STATE \texttt{ymin, ymax = clip(ys-yd,0,H), clip(ys+yd,0,H)} \\ 
			\RETURN{\texttt xmin,ymin,xmax,ymax}
		\end{algorithmic}
	\end{algorithm}

	\section{WSOL Results for Outer-Box}\label{apx-sec:wsol}
	
	In the main text, we have reported scores only using the inner-box connected component selection strategy based on the inner box.
	We present Tables \ref{tab:train-set-outer}, \ref{tab:combined-outer} and \ref{tab:combined-outer2} which extend Tables 5 and 6 with the results using both strategies.

	\begin{table}[!ht]
		\centering
		\caption{ViTS/16 AP@50\% performance on VOC07, VOC12, and COCO20k. Focus on the training dataset. Same as Table~5.a but also shows results for \emph{outer-box}.}% DINO and SLIC are reported for reference values.}
	\label{tab:train-set-outer}
	{   \small \centering
		\begin{tabular*}{\linewidth}{@{\extracolsep{\fill}}lccc}
			\toprule
			Training Dataset & VOC07 & VOC12 & COCO20k \\
			\midrule
			ImageNet1K            (inner-box) & 68.7 & 69.5 & 64.1 \\
			ImageNet1K            (outer-box) & 74.9 & 76.5 & 70.0 \\[1mm]
			Random Initialization (inner-box) & 45.2 & 46.9 & 48.0 \\
			Random Initialization (outer-box) & 52.9 & 54.2 & 57.2 \\[1mm]
			STL-10                (inner-box) & 44.2 & 44.7 & 51.3 \\
			STL-10                (outer-box) & 53.5 & 54.1 & 61.5 \\[1mm]
			Chest-X-RAY           (inner-box) & 42.8 & 45.1 & 42.7 \\
			Chest-X-RAY           (outer-box) & 49.4 & 51.6 & 50.9 \\
			\midrule \midrule
			DINO  (inner-box) & 62.7  & 62.3  & 61.0 \\
			DINO  (outer-box) & 74.3  & 76.1  & 68.4 \\[1mm]
			SLIC              & 43.1  & 46.0  & 38.5 \\
			\bottomrule
	\end{tabular*}}
\end{table}

\begin{table}[!ht]
	\centering
	\caption{AP@50\% performance on VOC07, VOC12, and COCO20k. We only depict results for NAVE. Same as Tables~5.b and c. but also shows results for \emph{outer-box}.
	}
	\label{tab:combined-outer}
	{   \small \centering
		\begin{tabular*}{\linewidth}{@{\extracolsep{\fill}}lccc}
			\toprule
			Feature            & VOC07 & VOC12 & COCO20k \\ \midrule
			\multicolumn{4}{c}{Training Scheme} \\
			ViTS/16 Random Init. (inner-box) & 45.2  & 46.9  & 48.0 \\ 
			ViTS/16 Random Init. (outer-box) & 52.9  & 54.2  & 57.2 \\[1mm]
			ViTS/16 ImageNet     (inner-box) & 68.7  & 69.5  & 64.1 \\ 
			ViTS/16 ImageNet     (outer-box) & 74.9  & 76.5  & 70.0 \\[1mm] 
			DINO-ViTS/16         (inner-box) & 63.1  & 64.3  & 60.6 \\ 
			DINO-ViTS/16         (outer-box) & 74.3  & 76.1  & 68.4 \\[1mm] 
			DINOv2-ViTS/14       (inner-box) & 66.0  & 66.2  & 65.1 \\ 
			DINOv2-ViTS/14       (outer-box) & 73.0  & 74.2  & 70.6 \\ 
			\midrule
			\multicolumn{4}{c}{Architecture} \\
			ViTS/16 ImageNet   (inner-box) & 68.7  & 69.5  & 64.1 \\ 
			ViTS/16 ImageNet   (outer-box) & 74.9  & 76.5  & 70.0 \\[1mm] 
			ViTB/16 ImageNet   (inner-box) & 44.1  & 45.3  & 48.6 \\ 
			ViTB/16 ImageNet   (outer-box) & 52.8  & 53.7  & 58.3 \\[1mm] 
			ResNet50 ImageNet  (inner-box) & 61.2  & 62.3  & 57.3 \\ 
			ResNet50 ImageNet  (outer-box) & 60.8  & 61.4  & 58.0 \\ 
			\bottomrule 
	\end{tabular*}}
\end{table}

\vfill 
\begin{table}[!ht]
	\centering
	\caption{AP@50\% performance on VOC07, VOC12, and COCO20k. We only depict results for NAVE. Same as Table~6 but also shows results for \emph{outer-box}.
	}
	\label{tab:combined-outer2}
	{   \small \centering
		\begin{tabular*}{\linewidth}{@{\extracolsep{\fill}}lccc}
			\toprule
			Feature            & VOC07 & VOC12 & COCO20k \\ \midrule
			\multicolumn{4}{c}{Registers} \\
			DINOv2-ViTS/14       (inner-box) & 66.0  & 66.2  & 65.1 \\ 
			DINOv2-ViTS/14       (outer-box) & 73.0  & 74.2  & 70.6 \\[1mm] 
			DINOv2-ViTS/14 + reg (inner-box) & 66.5  & 66.4  & 65.7 \\ 
			DINOv2-ViTS/14 + reg (outer-box) & 74.2  & 75.4  & 71.3 \\[1mm] 
			DINOv2-ViTB/14       (inner-box) & 63.8  & 64.0  & 63.5 \\ 
			DINOv2-ViTB/14       (outer-box) & 71.3  & 72.5  & 69.1 \\[1mm] 
			DINOv2-ViTB/14 + reg (inner-box) & 65.7  & 66.8  & 65.1 \\ 
			DINOv2-ViTB/14 + reg (outer-box) & 73.2  & 75.0  & 70.9 \\ 
			\bottomrule 
	\end{tabular*}}
\end{table}

\clearpage
\newpage

\section{WSOL Results for different $K$ values}\label{apx-sec:wsolK}

In Tables~\ref{apx-tab:combined-kmeans}, ~\ref{apx-tab:combined-kmeans2} and  \ref{apx-tab:combined-kmeans3}, we extend Tables 5 and 6 of the main paper for different numbers of clusters ($K=3$, $5$ or $7$).
We report here only values using the inner-box strategies. Overall, setting $K=5$ achieves the best performance.

\begin{table}[!ht]
	\centering
	\caption{AP@50\% performance on VOC07, VOC12, and COCO20k. We only depict results for NAVE (inner-box). Same as Table~5 (only $K=5$) but also shows results for $K=3$ and $K=7$. 
	}
	\label{apx-tab:combined-kmeans}
	{   \small \centering
		\begin{tabular*}{\linewidth}{@{\extracolsep{\fill}}lcccc}
			\toprule
			Feature            & $K$ & VOC07 & VOC12 & COCO20k \\ \midrule
			\multicolumn{5}{c}{Training Scheme} \\
			ViTS/16 Random Init.     & 3 & 52.6  & 49.2  & 49.4 \\ 
			ViTS/16 Random Init.     & 5 & 45.2  & 46.9  & 48.0 \\ 
			ViTS/16 Random Init.     & 7 & 41.7  & 41.3  & 45.5 \\[1mm] 
			ViTS/16 ImageNet         & 3 & 62.0  & 66.0  & 53.6 \\ 
			ViTS/16 ImageNet         & 5 & 68.7  & 69.5  & 64.1 \\ 
			ViTS/16 ImageNet         & 7 & 66.0  & 64.9  & 66.1 \\[1mm] 
			DINO-ViTS/16             & 3 & 58.6  & 62.8  & 51.2 \\ 
			DINO-ViTS/16             & 5 & 63.1  & 64.3  & 60.6 \\ 
			DINO-ViTS/16             & 7 & 58.5  & 55.2  & 62.3 \\[1mm] 
			DINOv2-ViTS/14           & 3 & 63.2  & 67.1  & 54.7 \\ 
			DINOv2-ViTS/14           & 5 & 66.0  & 66.2  & 65.1 \\ 
			DINOv2-ViTS/14           & 7 & 61.5  & 58.8  & 66.9 \\ 
			\bottomrule
	\end{tabular*}}
\end{table}

\begin{table}[!ht]
	\centering
	\caption{AP@50\% performance on VOC07, VOC12, and COCO20k. We only depict results for NAVE (inner-box). Same as Table~5 (only $K=5$) but also shows results for $K=3$ and $K=7$.
	}
	\label{apx-tab:combined-kmeans2}
	{   \small \centering
		\begin{tabular*}{\linewidth}{@{\extracolsep{\fill}}lcccc}
			\toprule
			Feature            & $K$ & VOC07 & VOC12 & COCO20k \\ \midrule
			\multicolumn{5}{c}{Architecture} \\
			%\midrule
			ViTS/16 ImageNet         & 3 & 62.0  & 66.0  & 53.6 \\ 
			ViTS/16 ImageNet         & 5 & 68.7  & 69.5  & 64.1 \\ 
			ViTS/16 ImageNet         & 7 & 66.0  & 64.9  & 66.1 \\[1mm] 
			ViTB/16 ImageNet         & 3 & 49.6  & 52.7  & 50.0 \\ 
			ViTB/16 ImageNet         & 5 & 44.1  & 45.3  & 48.6 \\ 
			ViTB/16 ImageNet         & 7 & 39.8  & 39.8  & 45.5 \\[1mm] 
			ResNet50 ImageNet        & 3 & 53.2  & 57.7  & 45.4 \\ 
			ResNet50 ImageNet        & 5 & 61.2  & 62.3  & 57.3 \\ 
			ResNet50 ImageNet        & 7 & 44.1  & 42.0  & 60.0 \\ 
			\bottomrule
	\end{tabular*}}
\end{table}

\newpage

\begin{table}[!ht]
	\centering
	\caption{AP@50\% performance on VOC07, VOC12, and COCO20k. We only depict results for NAVE (inner-box). Same as Table~6 (only $K=5$) but also shows results for $K=3$ and $K=7$. 
	}
	\label{apx-tab:combined-kmeans3}
	{   \small \centering
		\begin{tabular*}{\linewidth}{@{\extracolsep{\fill}}lcccc}
			\toprule
			Feature            & $K$ & VOC07 & VOC12 & COCO20k \\ \midrule
			\multicolumn{5}{c}{Training Scheme} \\
			\multicolumn{5}{c}{Registers} \\
			%\midrule
			DINOv2-ViTS/14           & 3 & 63.2  & 67.1  & 54.7 \\ 
			DINOv2-ViTS/14           & 5 & 66.0  & 66.2  & 65.1 \\ 
			DINOv2-ViTS/14           & 7 & 61.5  & 58.8  & 66.9 \\[1mm] 
			DINOv2-ViTS/14 + reg     & 3 & 65.2  & 68.6  & 56.2 \\ 
			DINOv2-ViTS/14 + reg     & 5 & 66.5  & 66.4  & 65.7 \\ 
			DINOv2-ViTS/14 + reg     & 7 & 61.6  & 58.7  & 67.2 \\[1mm] 
			DINOv2-ViTB/14           & 3 & 62.8  & 66.6  & 53.3 \\ 
			DINOv2-ViTB/14           & 5 & 63.8  & 64.0  & 63.5 \\ 
			DINOv2-ViTB/14           & 7 & 59.0  & 57.3  & 65.4 \\[1mm] 
			DINOv2-ViTB/14 + reg     & 3 & 64.8  & 68.6  & 55.7 \\ 
			DINOv2-ViTB/14 + reg     & 5 & 65.7  & 66.8  & 65.1 \\ 
			DINOv2-ViTB/14 + reg     & 7 & 61.3  & 58.7  & 67.0 \\ 
			\bottomrule
	\end{tabular*}}
\end{table}

\section{Extended Results for Unsupervised Segmentation}\label{apx-sec:useg}

\begin{figure}[!ht]
	\centering
	\includegraphics[width=\linewidth]{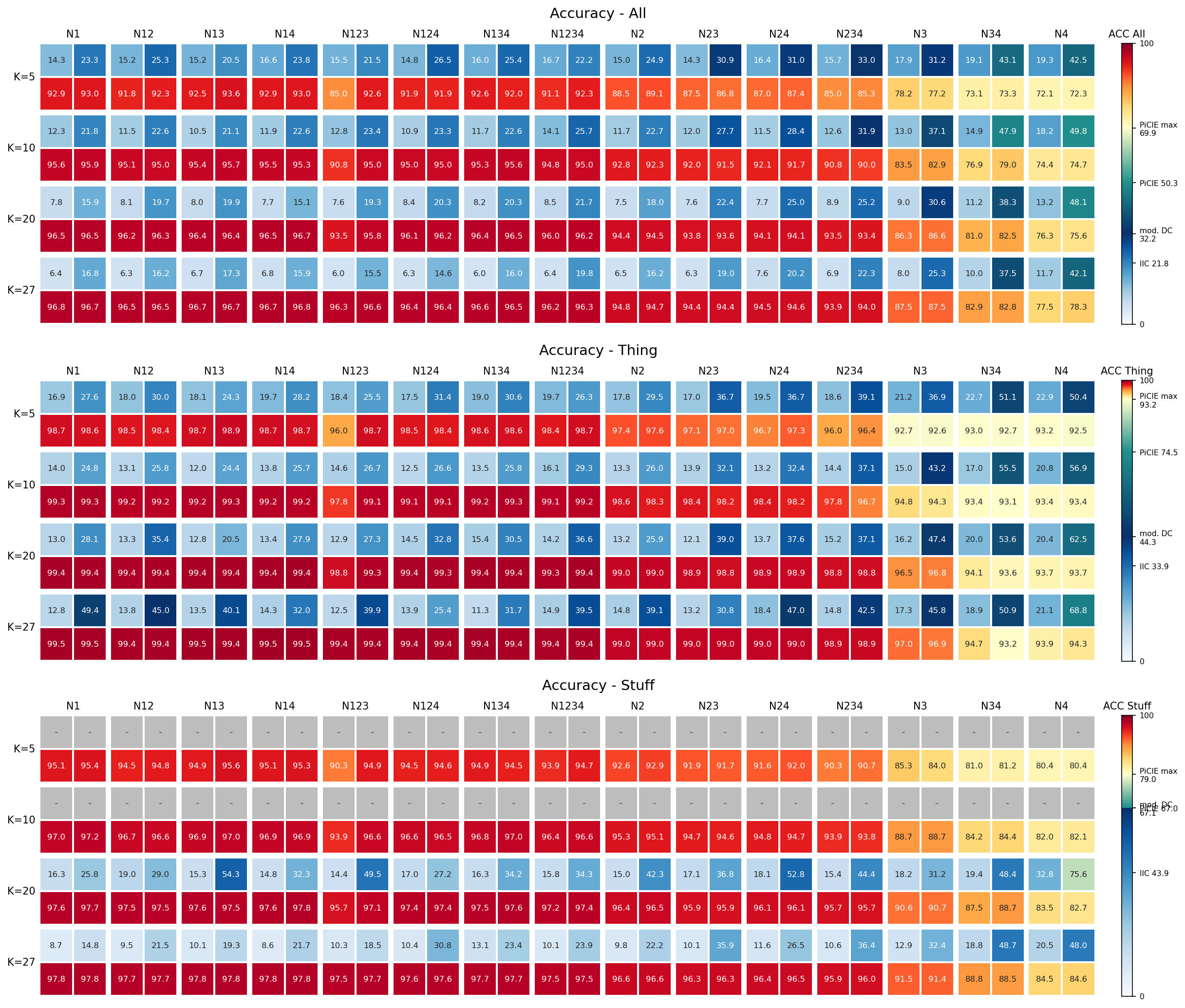}
	\caption{Ablation study of the number of clusters and selected layers for accuracy performance of NAVE on the unsupervised segmentation task.}
	\label{apx-fig:useg-acc}
\end{figure}

\begin{figure}[!ht]
	\centering
	\includegraphics[width=\linewidth]{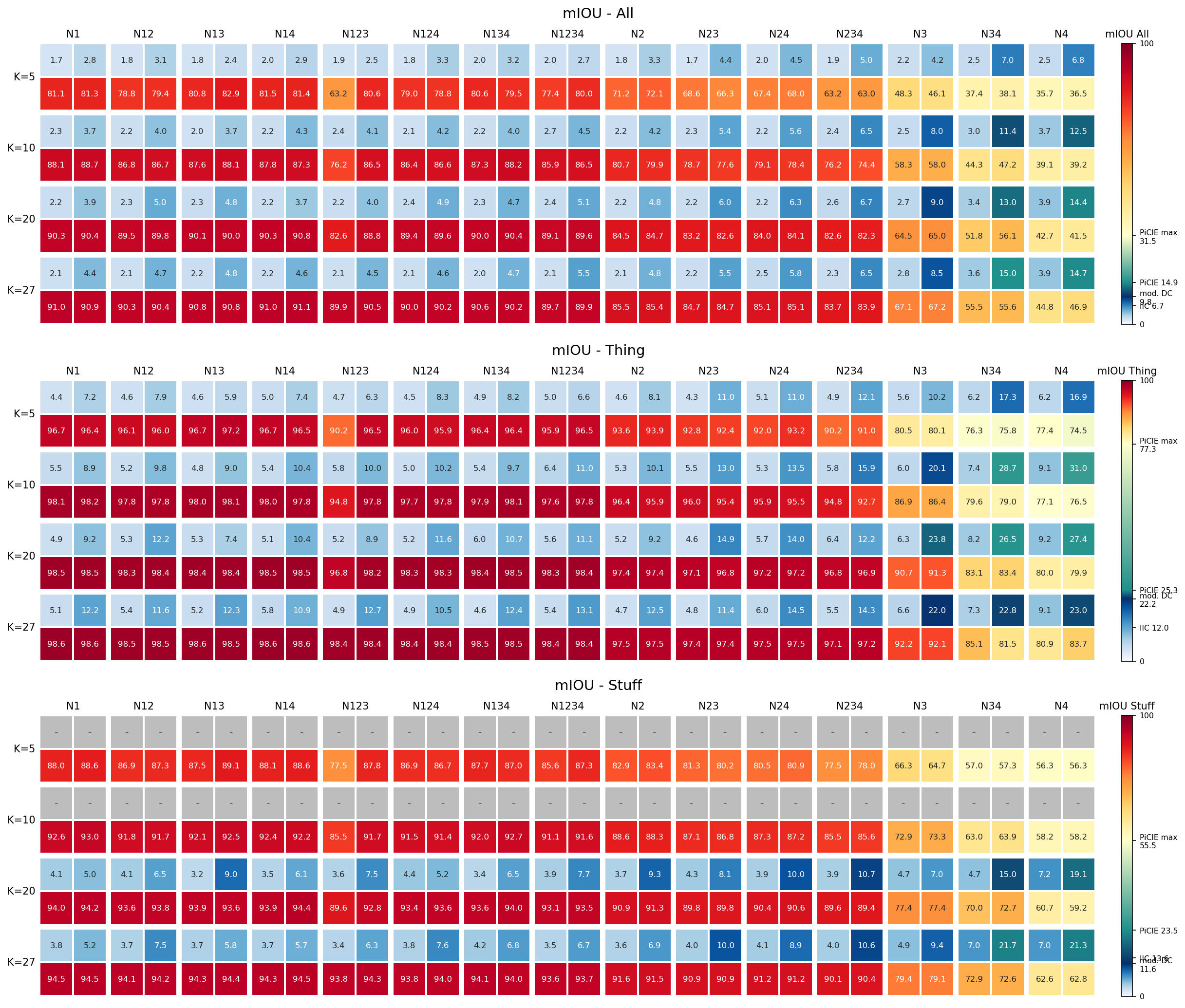}
	\caption{Ablation study of the number of clusters and selected layers for mean IoU performance of NAVE on the unsupervised segmentation task.}
	\label{apx-fig:useg-iou}
\end{figure}

In Figures~\ref{apx-fig:useg-acc} and~\ref{apx-fig:useg-iou}, we extend Table 5 of the main paper for different combinations of number of clusters and layers. Following [9], we also report performance on all the 27 super-classes of COCO20k and the Things and Stuff subsets. The colormap is built such that a score of 0 is white, the second best baseline is blue, the best baseline is green, PiCIE-max is yellow and the max score 100\% is red. Each cell is split in four configuration: The first row is without COCO-label maximizing reassignments, while the second is with it. For the first column, NAVE clustering is fitted on each batch, while the second column reports results when the clustering is fitted only with the first batch. Note that the test-set being not shuffled all runs are fitted on the same set. For reference, we also provide for six test images the ground truth annotation,the PiCIE segementation and NAVE explanations for $K=20$ and all layer combinations.

\begin{figure}[!ht]
	\centering
	\includegraphics[width=\linewidth]{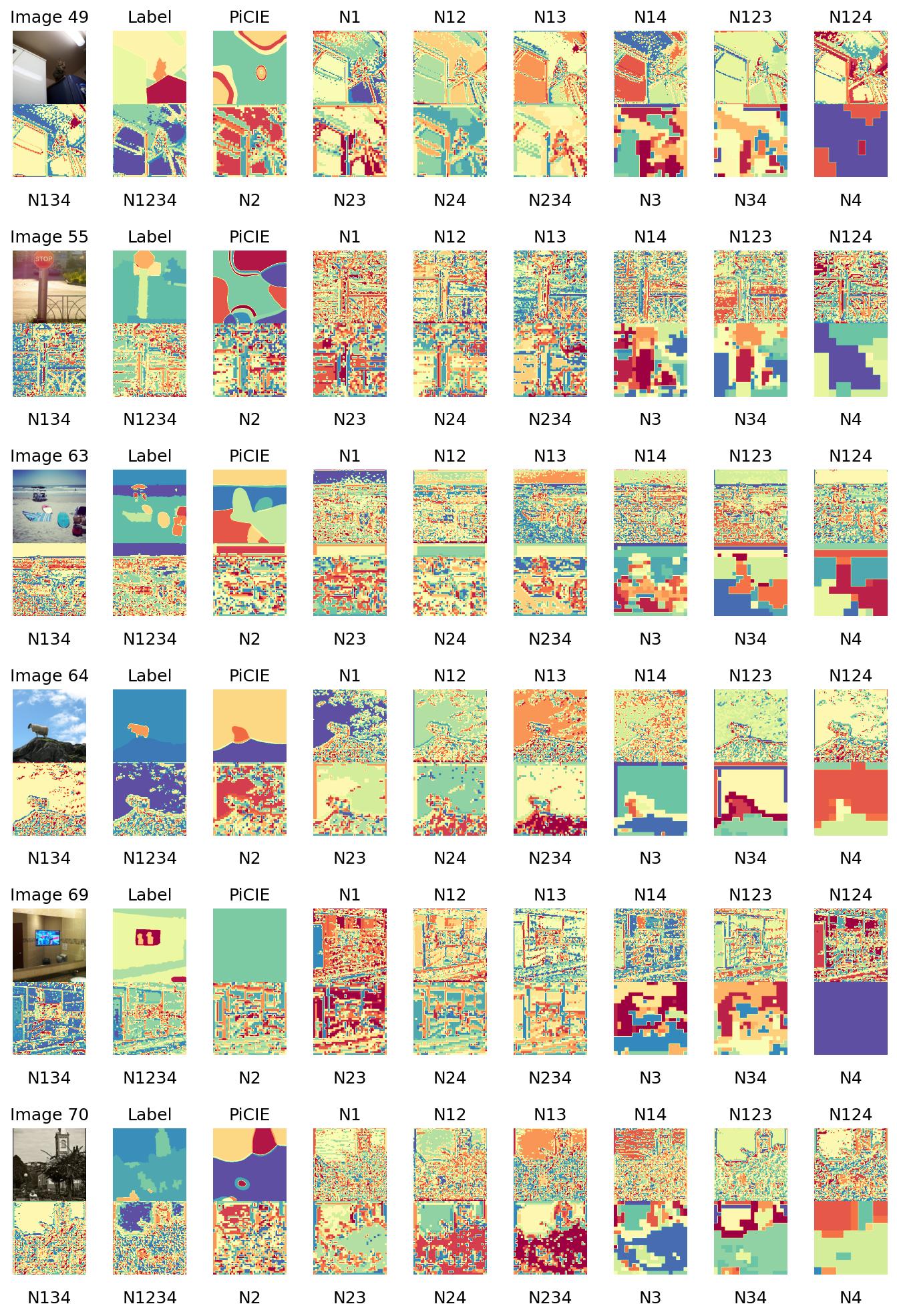}
	\caption{Examples of unsupervised segmentations produced by PiCIE and NAVE with different combination of layers and $K=20$.}
	\label{apx-fig:useg-expl}
\end{figure}

We observe that maximizing the concept assignments to the label yield almost 100\% accuracy and above 80\% mIoU when the output of the first residual block is involved. This is largely due to smaller segments (see Figure~\ref{apx-fig:useg-expl}). Nevertheless, even using only the output of the alst block (N4), the maximized performance are better than PiCIE and better or on par with PiCIE-max. 
Conversely, without concept-assignments, (first row of each cell), involving shallower layers harms the performance.
Surprisingly, fitting the clustering using only the first batch (second column) yields better performance than fitting a clustering on each batch. This behaviour call for further investigation.\\

\vfill

\vfill

\vfill

\vfill

\clearpage

\section{Explanation Maps for Different $K$ and Layers}\label{apx-sec:maps}

\begin{figure}[!ht]
	\centering
	\includegraphics[width=.78\linewidth]{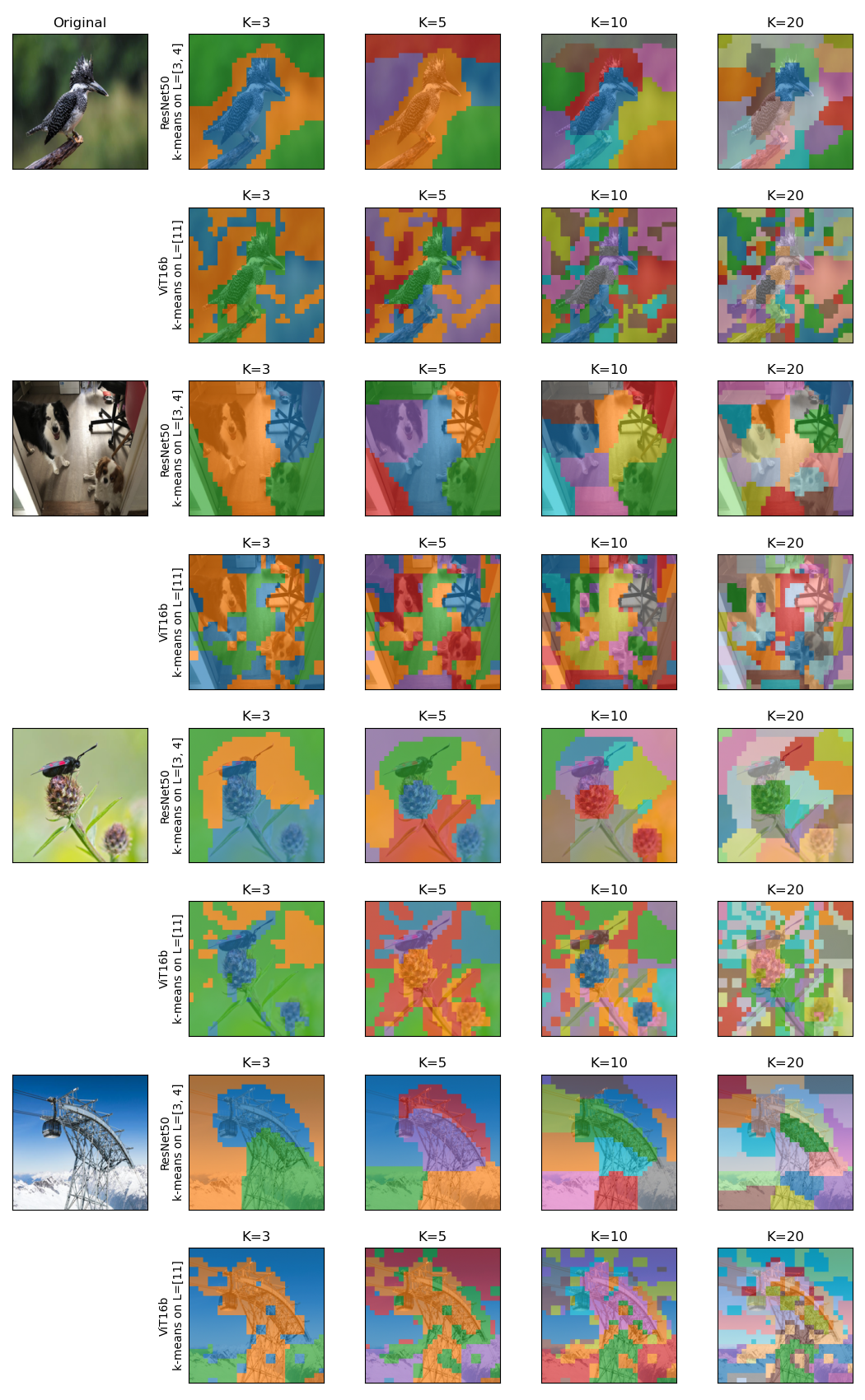}\\
	\caption{Ablation study of the number of clusters for NAVE for a ResNet50 and a ViTB16 pretrained on ImageNet1K.}
	\label{apx-fig:abl_k}
\end{figure}

\begin{figure*}
	\centering
	\includegraphics[width=\linewidth]{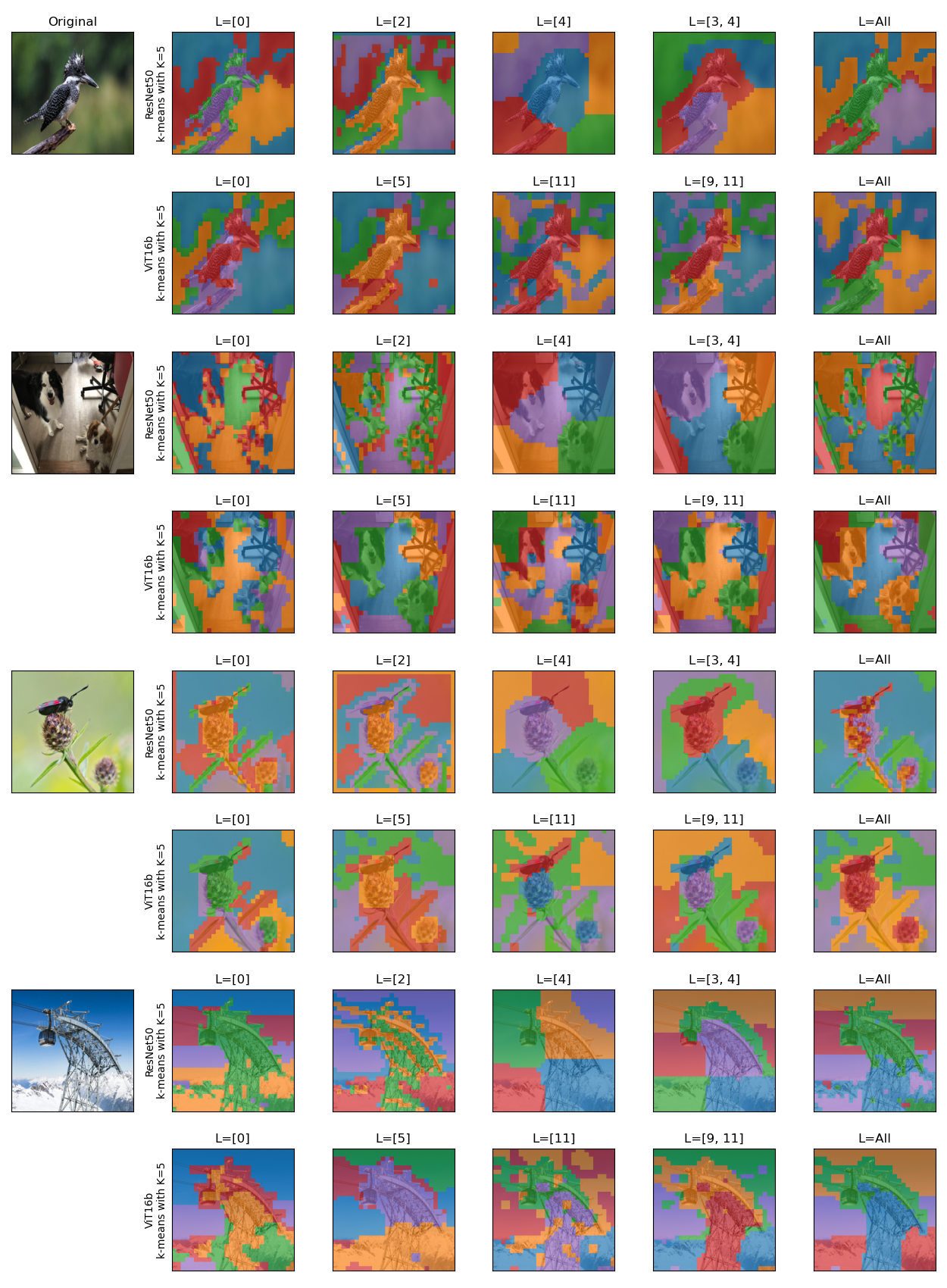}\\
	\caption{Ablation study of the layers used to compute NAVE for a ResNet50 and a ViTB16 pretrained on ImageNet1K.}
	\label{apx-fig:abl_l}
\end{figure*}

\begin{figure*}
	\centering
	\includegraphics[height=.93\textheight]{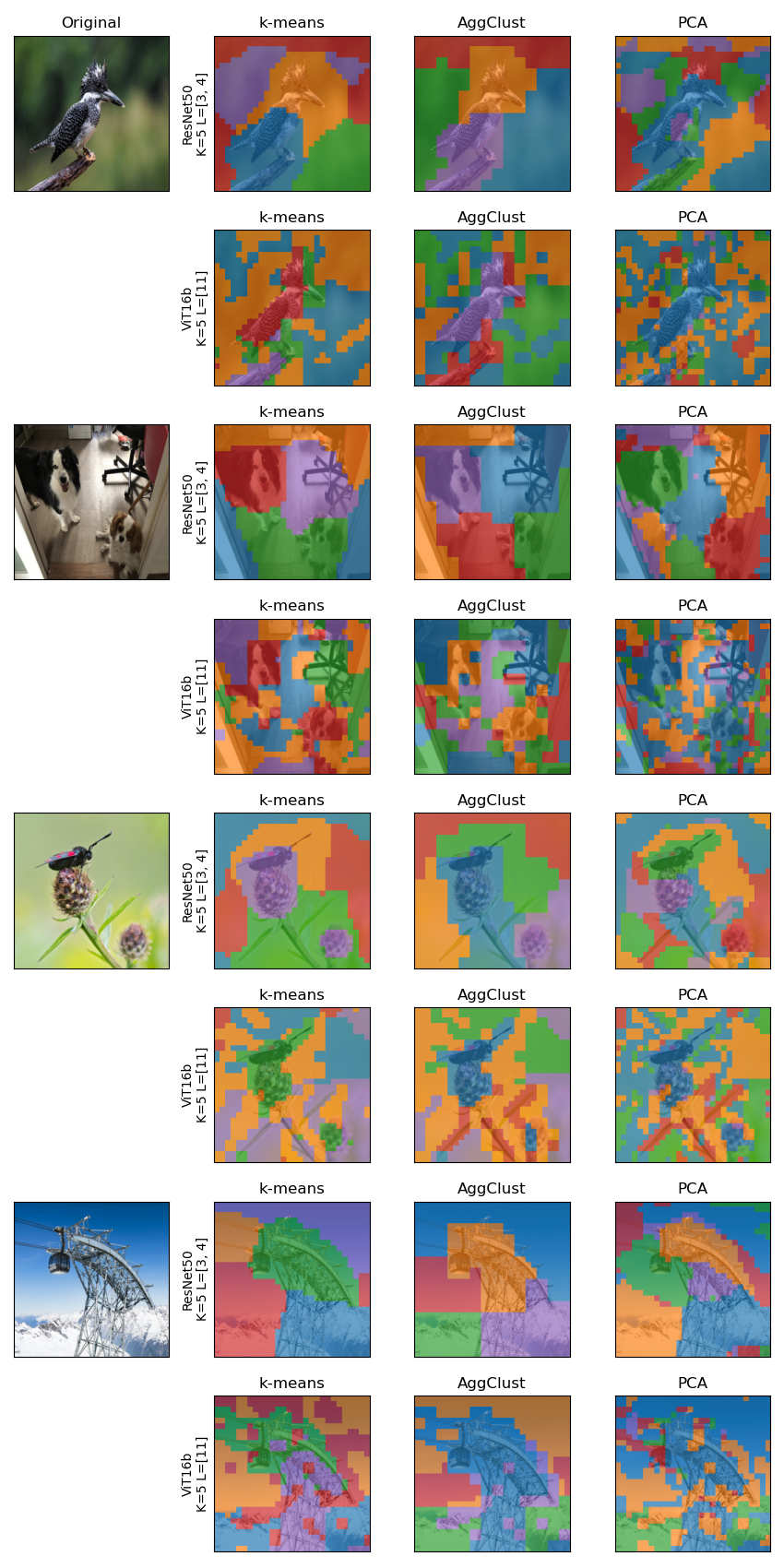}\\
	\caption{Ablation study of the clustering algorithm used in NAVE for a ResNet50 and a ViTB16 pretrained on ImageNet1K.}
	\label{apx-fig:abl_p}
\end{figure*}

The next three figures depict qualitative ablation studies for the number of clusters (Fig.~\ref{apx-fig:abl_k}), the extracted layers (Fig.~\ref{apx-fig:abl_l}), and the clustering algorithm (Fig.~\ref{apx-fig:abl_p}).\looseness=-1

\end{appendix}

\end{document}